%% file: main.tex
\documentclass[10pt,twocolumn,letterpaper]{article}

\usepackage[pagenumbers]{cvpr} 


\definecolor{cvprblue}{rgb}{0.21,0.49,0.74}
\usepackage[pagebackref,breaklinks,colorlinks,allcolors=cvprblue]{hyperref}

\definecolor{lightblue}{rgb}{0.8,0.85,1}
\usepackage{colortbl}
\usepackage[accsupp]{axessibility}  

\usepackage{multirow}

\def\paperID{2797} 
\def\confName{CVPR}
\def\confYear{2025}

\title{Distilling Long-tailed Datasets}

\author{
  Zhenghao Zhao$^{1}$\thanks{equal contribution}\qquad Haoxuan Wang$^{1}$\footnotemark[1]\qquad Yuzhang Shang$^{2}$\qquad Kai Wang$^{3}$\qquad Yan Yan$^{1}\thanks{corresponding author}$\\
  $^{1}$University of Illinois Chicago ~~
  $^{2}$Illinois Institute of Technology ~~
  $^{3}$National University of Singapore
\\
  {\tt\small \{zzhao72, hwang339, yyan55\}@uic.edu}
  ~~~~~ {\tt\small yshang4@hawk.iit.edu}
  ~~~~~ {\tt\small kai.wang@comp.nus.edu.sg}
}

\begin{document}
\maketitle
\input{sec/0_abstract}    
\input{sec/1_intro}
\input{sec/2_methods}
\input{sec/3_experiments}
\input{sec/4_related_work}
\input{sec/5_conclusion}

{
    \small
    \bibliographystyle{ieeenat_fullname}
    \bibliography{main}
}

\input{sec/X_suppl}


\end{document}

%% file: sec/0_abstract.tex
\begin{abstract}
    Dataset distillation aims to synthesize a small, information-rich dataset from a large one for efficient model training. However, existing dataset distillation methods struggle with long-tailed datasets, which are prevalent in real-world scenarios.
    By investigating the reasons behind this unexpected result, we identified two main causes: 
    1) The distillation process on imbalanced datasets develops biased gradients, leading to the synthesis of similarly imbalanced distilled datasets. 
    2) The experts trained on such datasets perform suboptimally on tail classes, resulting in misguided distillation supervision and poor-quality soft-label initialization.
    To address these issues, we first propose Distribution-agnostic Matching to avoid directly matching the biased expert trajectories. 
    It reduces the distance between the student and the biased expert trajectories and prevents the tail class bias from being distilled to the synthetic dataset. Moreover, we improve the distillation guidance with Expert Decoupling, which jointly matches the decoupled backbone and classifier to improve the tail class performance and initialize reliable soft labels. 
    This work pioneers the field of long-tailed dataset distillation, marking the first effective effort to distill long-tailed datasets. Our code will be made public at \url{https://github.com/ichbill/LTDD}.
\end{abstract}

%% file: sec/1_intro.tex
\section{Introduction}
\label{sec:intro}
Dataset distillation (DD) \cite{cazenavette2022datasetmtt,datm,dream} aims to synthesize small and high-quality data summaries which contain the most important knowledge from a given target dataset \cite{dd_survey}. 
Such a compact data summary offers numerous advantages, including faster model training, reduced memory consumption, and greater flexibility for various tasks \cite{dqas}, such as continual learning \cite{continual}, neural architecture search \cite{darts}, and knowledge distillation \cite{kd}. Various types of dataset distillation methods have been developed, including performance matching methods~\cite{loo2022efficientkernel1,nguyen2020datasetkernel3,zhou2022datasetkernel2}, parameter matching methods~\cite{datm,dream,zhao2020datasetdc}, representation matching methods~\cite{shang2024mim4dd,zhao2023improvedidm} and generative-based methods~\cite{gu2024efficientminimax, su2024d4m}. Currently, parameter matching methods have demonstrated significant performance improvements across various datasets.

\begin{figure}[t]
  \centering
    \centering
    \centerline{\includegraphics[trim=45 0 80 60,clip, width=1.0\linewidth]{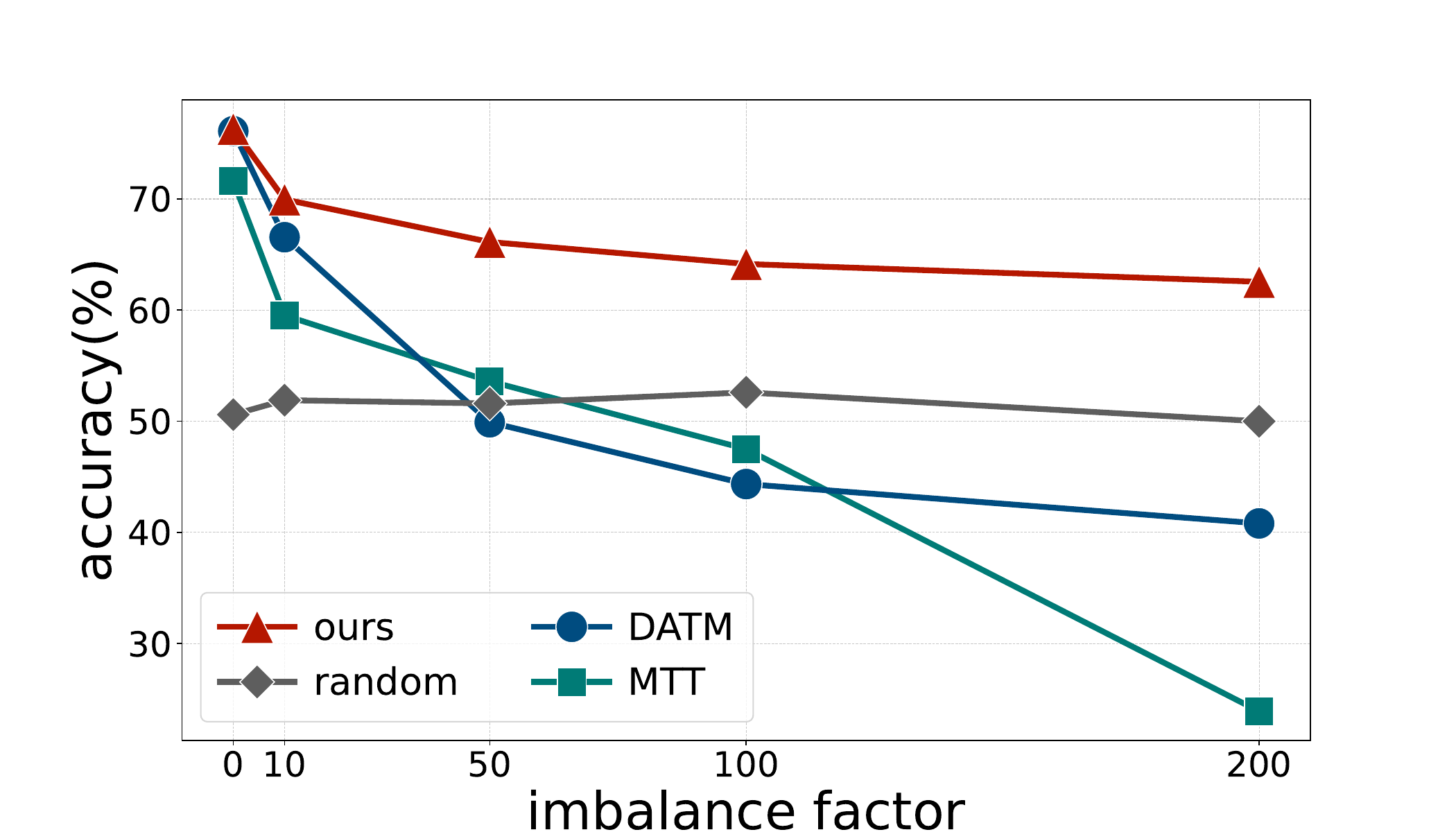}} 
    \vspace{-8pt}
    \caption{\textbf{Performance comparison on CIFAR-10-LT.} Existing Dataset Distillation methods exhibit degraded performance when applied to imbalanced datasets, especially when the imbalance factor increases, whereas our method provides significantly better performance under different imbalanced scenarios.}
    \label{fig_issue_1}
    \vspace{-19pt}
\end{figure}

Research on dataset distillation has predominantly focused on uniformly distributed datasets, such as CIFAR10, CIFAR100 \cite{cifar}, and TinyImageNet \cite{tiny-imagenet}. However, in real-world applications such as medical image diagnosis \cite{medical_lt}, training samples typically exhibit a long-tailed class distribution \cite{lt_survey}. In these scenarios, a small number of classes (head classes) have a large number of samples, while the majority of classes (tail classes) have only a few samples. 
Motivated by this discrepancy and the desire for practical application, we are interested in the novel task of long-tailed dataset distillation (LTDD). LTDD aims to distill a long-tailed target dataset into a small and uniformly distributed synthetic dataset, where models trained on this distilled dataset can achieve satisfactory performance on the uniformly distributed test set. 

\begin{figure}[t]
    \centering
    \centerline{\includegraphics[trim=10 0 0 0,clip,width=\linewidth]{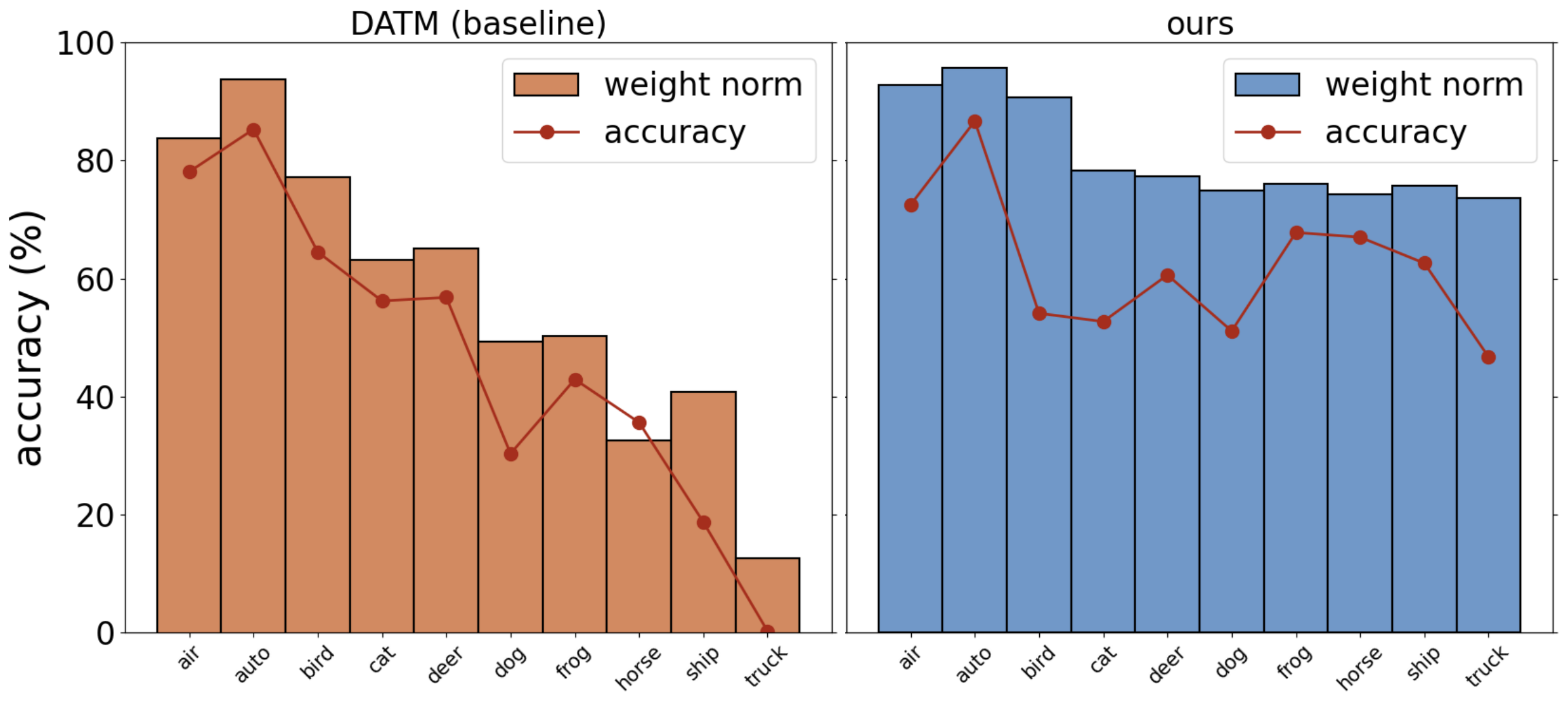}}
  \vspace{-8pt}
  \caption{\textbf{Relationship of classifier weights and class-wise accuracy.} We reveal that classifiers generated by existing dataset distillation methods often exhibit imbalanced distributions, resulting in poor performance on tail classes. In contrast, our method produces balanced classifiers, thereby enhancing overall accuracy.}
  \vspace{-20pt}
  \label{fig_issue_2}
\end{figure}

Nevertheless, distilling long-tailed datasets presents significant challenges, as the effectiveness of current dataset distillation methods diminishes when applied to such imbalanced distributions. As shown in Figure~\ref{fig_issue_1}, we find that the performance of different dataset distillation methods consistently decreases as the degree of imbalance increases.
The degree of imbalance is quantified by the imbalance factor, which is a commonly used metric to characterize the extent of imbalance \cite{krizhevsky2017imagenetconvnet}. The larger the imbalance factor is, the more imbalanced the dataset is.
When the imbalance factor is large, popular dataset distillation methods may even perform worse than random selection (e.g. DATM can only achieve 40.1\% accuracy when the imbalance factor $\beta=200$, while random selection can achieve 49.9\% accuracy). 
As shown in Figure \ref{fig_issue_2}, the performance degradation is concentrated in the tail classes. We attribute these performance failures to two key aspects of the learning procedure in long-tailed dataset distillation:

\noindent \textbf{(1)} During distillation, the weight distributions of the student and expert diverge, resulting in less informative tail class samples. 
Specifically, though the expert is trained on an imbalanced dataset, the student learns on a balanced one, resulting in differently distributed model weights. Such a discrepancy introduces controversy in the bi-level optimization objective, causing the distillation process to additionally focus on matching the data quantity distribution. Consequently, the tail classes of the distilled dataset contain significantly less useful information than the head classes, despite each class having the same number of samples. 

\noindent \textbf{(2)} The expert models yield suboptimal performance on the tail classes.
Concretely, the expert is biased toward the head classes when trained on imbalanced datasets \cite{ltr_regularization,ltr_calibration}, resulting in degraded tail class accuracy.
Since the student's performance is upper-bounded by that of the expert, the expressiveness of the distilled dataset is consequently restricted. Additionally, the image labels are obtained via the expert's prediction logits \cite{datm}. As the expert is biased toward the head classes, the prediction logits for the tail classes are, therefore, less reliable and confident. 

To address the aforementioned issues, we propose a trajectory-matching approach named \textbf{Distribution-agnostic Matching with Expert Decoupling}. Firstly, the student model is trained via a Distribution-agnostic Matching (DAM) strategy, which consists of a loss designed for matching experts trained on long-tailed data. Specifically, when training the student model on the balanced synthetic dataset, we mimic the procedure of training a model on an imbalanced dataset to reduce the distance between the student and the biased expert trajectories. 
Secondly, to mitigate the poor performance of the expert model and to enhance the overall distillation performance, an Expert Decoupling (ED) strategy is also proposed. In practice, we train representation experts and classification experts in a decoupled manner~\cite{kang2019decoupling} to improve the performance of experts. We jointly match with the backbone layers of the representation experts and the classifier layers of the classification experts simultaneously. We use classification expert models to generate soft labels with high confidence for all classes. 

This work pioneers the field of long-tailed dataset distillation as the first successful effort in this area. Our experiments on four long-tailed datasets, conducted under various imbalance factors and images-per-class (IPC) settings, demonstrate the effectiveness of our method. We show that our method is able to achieve state-of-the-art performance across different dataset distillation baselines. 

%% file: sec/2_methods.tex
\section{Long-tailed Dataset Distillation}
\label{sec:method}

\subsection{Problem Formulation}
\label{sec:preliminaries}

\begin{figure*}[t]
  \centering
  \begin{subfigure}{0.32\linewidth}
    \centering
    \captionsetup{justification=centering}
    \centerline{\includegraphics[trim=0 20 360 0,clip, width=1.0\textwidth]{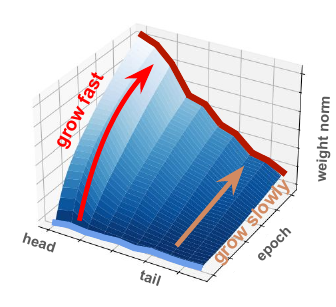}} 
    \vspace{-5pt}
    \caption{expert trajectory}
    \label{fig_expert_issue_1}
  \end{subfigure}
  \hfill
  \begin{subfigure}{0.32\linewidth}
    \centering
    \captionsetup{justification=centering}
    \centerline{\includegraphics[trim=181 20 181 0,clip, width=1.0\textwidth]{assets/grad_issue_3.pdf}} 
    \vspace{-5pt}
    \caption{training trajectory of traditional DD}
    \label{fig_expert_issue_2}
  \end{subfigure}
  \hfill
  \begin{subfigure}{0.32\linewidth}
    \centering
    \captionsetup{justification=centering}
    \centerline{\includegraphics[trim=360 20 0 0,clip, width=1.0\textwidth]{assets/grad_issue_3.pdf}} 
    \vspace{-5pt}
    \caption{training trajectory of \textbf{our method}}
    \label{fig_expert_issue_3}
  \end{subfigure}
  \hfill
  \vspace{-8pt}
  \caption{\textbf{Effect of biased expert.} (a) An expert trained on an imbalanced dataset leads to increasingly imbalanced weight gradients over classes. (b) Existing dataset distillation methods \cite{cazenavette2022datasetmtt,datm} are ignorant of the distribution gap between $\mathcal{S}_{t}$ and $\mathcal{D}$. This causes the student gradient imbalance to increase in each step. If we match such trajectories, the synthetic dataset will be updated by the increasingly imbalanced gradients over classes, and the model trained on this synthetic dataset is highly biased. (c) With Distribution-agnostic Matching, the increasingly imbalanced gradients over classes will be re-weighted, such that the student model is updated with balanced gradients.}
  \vspace{-20pt}
  \label{fig:expert_issue}
\end{figure*}

Assume we are given a long-tailed target training dataset $\mathcal{D}=\{(\mathbf{x}_i, y_i)\}^{|\mathcal{D}|}_{i=1}$ with $C$ classes, where $y \in \{0, \dots, C-1\}$. Denote the subset of images belonging to class $c$ as $\mathcal{D}_c$, then $|\mathcal{D}_{0}| > |\mathcal{D}_{1}| > \cdot \cdot \cdot > |\mathcal{D}_{C-1}|$ and $|\mathcal{D}_{0}| \gg |\mathcal{D}_{C-1}|$. 
Denote the test dataset as $\mathcal{T} = \{(\mathbf{x}_i, y_i)\}^{|\mathcal{T}|}_{i=1}$ and is balanced. 
Our goal is to generate a small dataset $\mathcal{S}=\{(\mathbf{\hat{x}}_i, \hat{y}_i)\}^{|\mathcal{S}|}_{i=1}$, where $|\mathcal{S}| \ll |\mathcal{D}|$ and $|\mathcal{S}_{0}| = |\mathcal{S}_{1}| = \cdot \cdot \cdot = |\mathcal{S}_{C-1}|$, so that models trained on $\mathcal{S}$ perform optimally on the test set $\mathcal{T}$.

Typical dataset distillation methods~\cite{cazenavette2022datasetmtt, dream, datm, cui2023scalingtesla} use the following procedure to optimize $\mathcal{S}$: First, train an expert model $\mathcal{M}_{\mathcal{D}}$ on $\mathcal{D}$ with acceptable performance. Then initialize $\mathcal{S}$ by randomly selecting images from $\mathcal{D}$ or with Gaussian noise, and predict the corresponding (soft) labels using $\mathcal{M}_{\mathcal{D}}$. A student model $\mathcal{M}_{\mathcal{S}}$ is then trained based on $\mathcal{S}$, and the distilled image set $\mathcal{S}$ is optimized by minimizing:
\begin{equation}
    \label{eq:distill}
     \mathcal{L}(\mathcal{S}) = || \mathcal{F}(\mathcal{M}_{\mathcal{D}}; \mathcal{D}) - \mathcal{F}(\mathcal{M}_{\mathcal{S}}; \mathcal{S}) ||_{p}
\end{equation}
where $\mathcal{F}$ are model characteristics that can be used for matching, such as feature distributions \cite{cafe}, gradients \cite{idc}, or training trajectories \cite{cazenavette2022datasetmtt,datm}. Trajectory matching~\cite{cazenavette2022datasetmtt, du2023minimizing, datm} 
matches the student model trained on the distilled dataset with the expert model trajectories trained on the original dataset, providing an impressive performance, 
and the matching loss is formulated as:
\begin{equation}
    \label{eq:tr_loss}
    \mathcal{L}_{\text{match}}(\hat{\theta}_{t+N},\theta^{*}_{t+M},\theta^{*}_{t}) = \frac{\|\hat{\theta}_{t+N}-\theta^*_{t+M}\|_2^2}{\|\theta^*_{t}-\theta^*_{t+M}\|_2^2},
\end{equation}
where $\theta^*_{t}$ and $\theta^*_{t+M}$ are the expert parameters on epoch $t$ and $t+M$. $\hat{\theta}_{t+N}$ is the student parameter obtained in the inner optimization updated using the cross-entropy loss.

In Section \ref{sec:dam}, we first discuss why such a formulation in Equation \ref{eq:distill} is unreasonable for a long-tailed target dataset. 
Then we propose Distribution-agnostic Matching to mitigate the negative impact of the long-tailed distribution. To further enhance the distillation performance, an Expert Decoupling strategy is proposed in Section \ref{sec:ed}, aiding our method to achieve improved and even lossless performance.

\begin{figure*}[t]
    \centering
    \centerline{\includegraphics[trim=120 70 120 80,clip, width=0.99\textwidth]{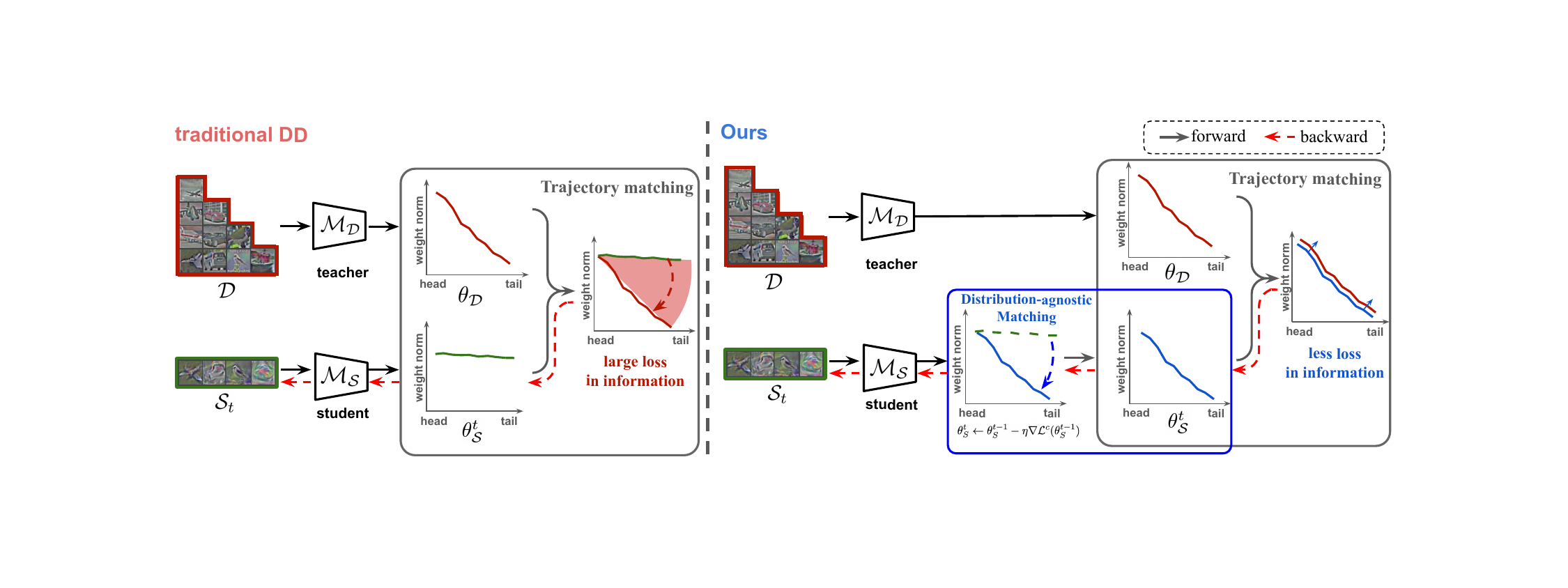}}
    \vspace{-0.25in}
    \caption{
    \textbf{Comparison of the internal loop between traditional DD methods and ours.} The expert model $\mathcal{M_{\mathcal{D}}}$ is trained on a long-tailed dataset, leading to biased weights. \textcolor{red}{Left:} For traditional DD methods, directly matching with these weights causes large information loss in the tail classes. The backward propagation updates this imbalance from the expert trajectories to the synthetic dataset. \textcolor{blue}{Right:} With Distribution-agnostic Matching, the gradients are obtained via $\mathcal{L}^c$, which revises the student weight for matching in the internal loop. This mitigates the distance between the student weight and the expert trajectory to reduce the influence of imbalance on the synthetic dataset.}
    \label{fig:swl_structure}
    \vspace{-16pt}
\end{figure*}

\subsection{Distribution-agnostic Matching}
\label{sec:dam}
\textbf{Influence of long-tailed distribution.}
By extending Equation \ref{eq:distill} to the whole distillation process, a single step of the synthetic dataset optimization can be formulated as:
\begin{align}
    \label{eq:eq2}
    \mathcal{S}_{t+1} = \underset{\mathcal{S}_{t}}{\arg\min} ||\theta_{\mathcal{D}} - \underset{\theta_{S}^{t}}{\arg\min}\mathcal{L}(\theta_\mathcal{S}^{t}, \mathcal{S}_{t})||_{p},
\end{align}
where $\theta_{\mathcal{S}}^{t}$ is the student model parameter at epoch $t$ and $\mathcal{L}$ is the cross-entropy loss based on the student model parameter and the synthetic dataset $\mathcal{S}_{t}$. $\theta_{\mathcal{D}}$ is the expert model parameters trained on $\mathcal{D}$ and is determined before distillation. $p$ is the factor for $L_{p}$ norm and usually takes the value of 2.

Equation \ref{eq:eq2} is an embedded optimization problem in which the inner loop is performed first. This implies that $\mathcal{S}_{t}$ is updated to train a $\theta_{\mathcal{S}}^{t}$ that can match the fixed $\theta_{\mathcal{D}}$ as much as possible. We see that $\theta_{\mathcal{S}}^{t}$ is an important intermediate that connects the synthetic dataset and the target dataset.
Under the setting that the target dataset $\mathcal{D}$ is imbalanced, the weights in $\theta_{\mathcal{D}}$ for different classes follow an imbalanced distribution and are fixed during distillation. Then the outer loop in Equation \ref{eq:eq2} would force $\theta_{\mathcal{S}}^{t}$ to produce an imbalanced distribution similar to $\theta_{\mathcal{D}}$. As illustrated in Figure~\ref{fig:expert_issue}(a) and (b), the student is forced to mimic the expert during distillation, so that its weight distribution in different classes gradually becomes imbalanced.

However, the outer loop in Equation \ref{eq:eq2} aims to find a valid $\mathcal{S}_{t}$ rather than a proper $\theta_{\mathcal{S}}^{t}$. Thus, $\mathcal{S}_{t}$ is optimized to resemble the training behavior of $\mathcal{D}$. Although the tail class performance on the long-tailed dataset $\mathcal{D}$ is poor, this causes the tail classes of distilled $\mathcal{S}_{t}$ to similarly contain much less useful information than the head classes, even though it is uniformly distributed. As a result, Equation \ref{eq:eq2} proactively distills less information from tail classes, making the distillation process more biased towards head classes.

\noindent\textbf{Distribution-agnostic Matching.}
As discussed above, matching the student network with an expert trained on an imbalanced dataset could lead to severe flaws in the synthetic dataset $\mathcal{S}$, where the tail classes are unfortunately distorted. Ideally, we would like $\mathcal{S}$ to be a compact data summary such that every distilled image is meaningful and useful. However, such a dataset yields a student model with a balanced distribution (as shown in Figure~\ref{fig:swl_structure} left), which mismatches the expert. This raises an interesting question: how can we preserve tail class information in the distilled dataset while avoiding the weight mismatching?

\textbf{A key insight is to weaken the relationship between the data distribution and the weight distribution}. In other words, we aim to prevent weight distribution mismatches from impacting dataset optimization. To achieve this, we propose Distribution-agnostic Matching, which seeks to align the gradient distributions of the student and expert without altering the distilled dataset. Practically, we modify the student training procedure to absorb the effect of weight imbalance by deliberately making $\theta_\mathcal{S}^{t}$ imbalanced. Inspired by the Balanced Softmax loss~\cite{ren2020balancedsoftmaxloss}, we propose a long-tailed distributed loss to aid student training:
\begin{equation}
    \label{eq:eq3}
    \mathcal{L}^{\text{c}}(\mathbf{L}, \mathbf{y}, \mathbf{s}) = -\frac{1}{n} \sum_{i=1}^{n} g(\mathbf{s}_{\mathbf{y}_i})\cdot\log P(\mathbf{L}_{i,\mathbf{y}_i}),
\end{equation}
where 
\begin{equation}
    \label{eq:eq3_1}
    P(\mathbf{L}_{i,\mathbf{y}_i}) = \frac{e^{\mathbf{L}_{i, \mathbf{y}_i} - \lambda\log(\mathbf{s}_{\mathbf{y}_i})}}{\sum_{j=1}^{C} e^{\mathbf{L}_{i,j} - \lambda\log(\mathbf{s}_j)}}.
\end{equation}
In the above equations, $\mathbf{L}$ is the student model logits, $n$ is the number of samples in the distilled dataset, $\mathbf{y}$ is the ground-truth label, $\mathbf{s}_{\mathbf{y}_i}$ is the number of samples in the long-tailed dataset for class $\mathbf{y}_{i}$, $g(\cdot)$ normalizes $\mathbf{s}_{\mathbf{y}_i}$, and $\lambda$ is used for smoothing the predictions.

The proposed loss is designed to support student model training during distillation by reducing the dependency of the weight distribution on the data distribution, thereby “mimicking” the training behavior on long-tailed datasets. Classes with more samples contribute more significantly to the loss, simulating a long-tailed training gradient for the student network during matching. This approach helps to minimize the trajectory differences between the student and expert networks caused by distribution mismatches between the target and distilled datasets. The right plot in Figure \ref{fig:swl_structure} illustrates the design and effect of our method.

\subsection{Improved Guidance with Expert Decoupling}
\label{sec:ed}

\begin{table*}[t]
\centering
\resizebox{1\linewidth}{!}{
\begin{tabular}{l|ccc|ccc|ccc|ccc}
\toprule
Dataset & \multicolumn{12}{c}{CIFAR-10-LT} \\
\midrule
Imbalance Factor & \multicolumn{3}{c|}{10} & \multicolumn{3}{c|}{50} & \multicolumn{3}{c|}{100} & \multicolumn{3}{c}{200}\\
IPC & 10 & 20 & 50 & 10 & 20 & 50 & 10 & 20 & 50 & 10 & 20 & 50 \\
\midrule
Random &32.5$\pm$2.2 & 39.6$\pm$0.9 & 51.9$\pm$1.5 & 33.2$\pm$0.4 & 42.0$\pm$1.3 & 51.6$\pm$1.3 & 34.4$\pm$2.0 & 41.4$\pm$0.7 & 52.6$\pm$0.5 & 32.5$\pm$0.8 & 42.2$\pm$1.1 & 49.9$\pm$1.4 \\
K-Center Greedy~\cite{sener2017activekcenter} & 21.9$\pm$0.8 & 24.2$\pm$0.8 & 31.7$\pm$0.9 & 17.8$\pm$0.2 & 20.8$\pm$0.5 & 26.1$\pm$0.2 & 16.2$\pm$0.5 & 19.0$\pm$1.0 & 24.2$\pm$1.2 & 16.8$\pm$0.3 & 17.5$\pm$1.4  & 22.6$\pm$1.6 \\
Graph-Cut~\cite{iyer2021submodulargraphcut} & 28.7$\pm$0.9 & 34.2$\pm$1.0 & 40.6$\pm$1.0 & 24.2$\pm$0.7 & 28.6$\pm$0.8 & 33.9$\pm$0.4 & 22.9$\pm$0.9 & 26.0$\pm$0.5 & 33.3$\pm$1.0 & 22.3$\pm$0.9 & 25.2$\pm$0.5 & 29.2$\pm$0.4 \\
\midrule
DC~\cite{zhao2020datasetdc} & 37.9$\pm$0.9 & 38.5$\pm$0.9 & 37.4$\pm$1.4 & 37.3$\pm$0.9 & 38.8$\pm$1.0 & 35.8$\pm$1.2 & 36.7$\pm$0.8 & 38.1$\pm$1.0 & 35.3$\pm$1.4 & 35.6$\pm$0.8 & 35.7$\pm$0.9 & 33.3$\pm$1.4 \\
MTT~\cite{cazenavette2022datasetmtt} & 58.0$\pm$0.8 & 59.5$\pm$0.4 & 62.0$\pm$0.9 & 45.8$\pm$1.4 & 49.9$\pm$0.8 & 53.6$\pm$0.5 & 37.7$\pm$0.6 & 41.6$\pm$1.1 & 47.8$\pm$1.1 & N/A & 22.6$\pm$1.0 & 23.9$\pm$0.8 \\
DREAM~\cite{dream} & 34.6$\pm$0.6 & 42.2$\pm$1.5 & 50.5$\pm$0.7 & 30.8$\pm$0.6 & 38.4$\pm$0.3 & 45.5$\pm$0.9 & 30.8$\pm$1.7 & 34.9$\pm$0.8 &42.2$\pm$0.8 & 32.7$\pm$1.3 & 32.4$\pm$0.3 & 38.9$\pm$0.4 \\
IDM~\cite{zhao2023improvedidm} & 54.8$\pm$0.4 & 57.1$\pm$0.3 & 60.1$\pm$0.3 & 51.9$\pm$0.7 & 53.3$\pm$0.6 & 56.1$\pm$0.4 & 49.8$\pm$0.6 & 50.9$\pm$0.5 & 53.1$\pm$0.4 & 47.0$\pm$0.5 & 48.1$\pm$0.5 & 49.9$\pm$0.3 \\
Minimax~\cite{gu2024efficientminimax} & 29.2$\pm$0.5 & 28.5$\pm$0.6 & 39.9$\pm$0.1 & 18.4$\pm$0.3 & 22.5$\pm$0.2 & 25.2$\pm$0.2 & 19.9$\pm$0.4 & 23.3$\pm$0.2 & 28.0$\pm$0.6 & 19.1$\pm$0.4 & 20.5$\pm$0.2 & 22.7$\pm$0.3 \\
DATM~\cite{datm} & 57.2$\pm$0.4 & 60.4$\pm$0.2 & 66.7$\pm$0.6 & 41.6$\pm$0.2 & 43.4$\pm$0.3 & 50.3$\pm$0.2 & 37.3$\pm$0.2 & 38.9$\pm$0.1 & 44.3$\pm$0.1 & N/A & 34.8$\pm$0.1 & 40.1$\pm$0.2 \\
\rowcolor{lightblue}\textbf{Ours} & \textbf{58.1$\pm$0.3} & \textbf{63.0$\pm$1.0} & \textbf{70.5$\pm$0.4} & \textbf{54.2$\pm$1.0} & \textbf{59.4$\pm$0.7} & \textbf{65.8$\pm$0.2} & \textbf{53.4$\pm$0.1} & \textbf{58.2$\pm$0.6} & \textbf{64.0$\pm$0.9} & \textbf{52.2$\pm$0.6} & \textbf{56.6$\pm$0.4} & \textbf{62.3$\pm$0.3} \\
\midrule
Full Dataset & \multicolumn{3}{c|}{76.7$\pm$0.3} & \multicolumn{3}{c|}{69.8$\pm$0.3} & \multicolumn{3}{c|}{66.2$\pm$0.4} & \multicolumn{3}{c}{62.1$\pm$0.4} \\
\bottomrule[1pt]
\end{tabular}
}
\vspace{-0.3cm}
\caption{\textbf{Quantitative comparisons with the SOTA methods on CIFAR-10-LT.} Our method outperforms all existing approaches, and can even achieve lossless performance under certain settings.
Performance is evaluated under various imbalance factors. As the first to explore long-tailed dataset distillation, all experiments are conducted using open-source code. N/A indicates distillation failure.
}
\label{tab:cifar10}
  \vspace{-18pt}
\end{table*}

Unlike experts trained on balanced target datasets, the experts trained on long-tailed datasets are less effective for guiding the dataset distillation process. Specifically, when adopting typical dataset distillation methods, the long-tailed experts perform poorly on tail classes, degrading student performance on these classes and leading to unreliable initial soft labels. This degraded student performance restricts the effective learning of synthetic tail-class images, while the unreliable labels introduce misleading information and provide weak supervision.

\begin{figure}[t]
    \centering
    \includegraphics[trim=0 200 0 200,clip, width=1.0\linewidth]{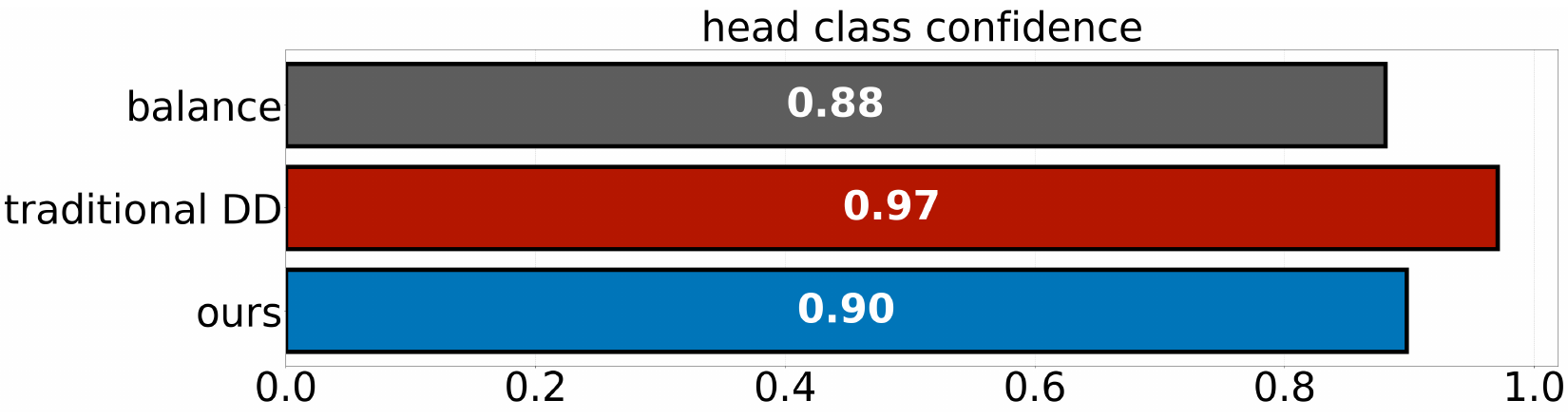}
    \includegraphics[trim=0 200 0 200,clip, width=1.0\linewidth]{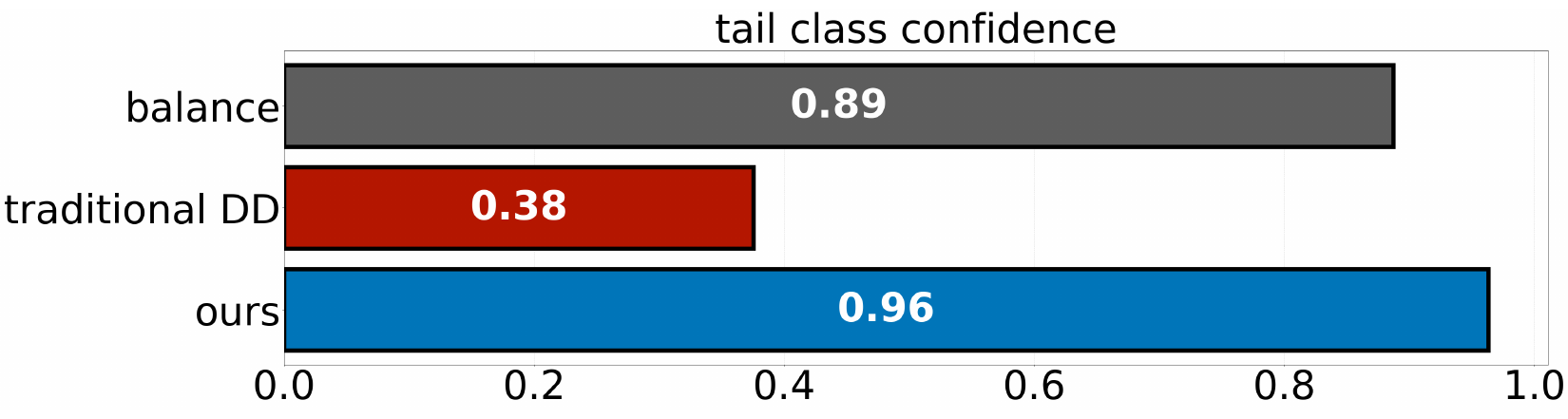} 
    \vspace{-20pt}
    \caption{\textbf{Soft-label initialization on CIFAR-10-LT.} We visualize the average predictive confidence of experts on different classes. For traditional DD, while the soft labels can be initialized well on head classes, they are predicted poorly on the tailed classes, leading to insufficient supervision.}
    \label{fig:ddd_moti}
    \vspace{-16pt}
\end{figure}

In the conventional dataset distillation process, soft labels are assigned to initialized synthetic images based on the probability output of the expert model~\cite{datm}. We found that soft-label predictions from long-tailed experts are distributed differently compared to those from balanced experts.
As visualized in Figure~\ref{fig:ddd_moti}, while the head class soft labels predicted by the balanced expert have similar confidence values with tail classes (0.88 and 0.89), the long-tailed expert predictions between head and tail classes differ drastically. Whereas the head class confidence value can be as large as 0.97, the confidence value of the tail class is surprisingly small (0.38). Such a confidence discrepancy leads to less effective learning in tail classes than in head classes. Moreover, for tail classes with a low number of samples, the number of samples with high confidence is reduced and leads to the diversity decrease of the dataset initialization.

Decoupled training~\cite{kang2019decoupling} is a widely used technique to alleviate the prediction bias introduced by imbalanced data. It consists of two stages: representation learning and classifier fine-tuning. At the representation learning stage, the entire model is trained on the original dataset $\mathcal{D}$. At the classifier fine-tuning stage, the backbone of the model is frozen, and only the classifier of the model is trained on a balanced dataset $\mathcal{B}$, which is created from $\mathcal{D}$ via oversampling or undersampling. However, the integration of this decoupling approach with dataset distillation is not trivial, as it requires careful alignment of the student model with the expert model. To effectively utilize the idea of Expert Decoupling, we propose a two-step strategy to improve teacher guidance for student training.

Firstly, we train two types of experts in a decoupled manner. The representation expert with parameter $\theta^\mathcal{D}$ is obtained from the representation learning stage, and the classifier expert with parameter $\theta^\mathcal{B}$ is obtained from the classifier finetuning stage. 
Since the classifier experts are not biased and are able to provide more accurate classification, we use the classifier experts $\theta^\mathcal{B}$ for the synthetic dataset $\mathcal{S}$ initialization. Specifically, we use the logits $L_i = \theta^\mathcal{B}\mathbf{x}_i$ to initialize soft labels, with $\text{softmax}(L_i)$ as the label of $\mathbf{x}_i$, and then sample from $\mathcal{D}$ to initialize $\mathcal{S}$.

Secondly, we match the teacher expert's training trajectory with the student's in an adaptive manner. Since the representation expert learns the image features from the original dataset $\mathcal{D}$ but fails to classify with biased classifiers~\cite{kang2019decoupling}, we skip matching the classifier layer when matching the student network trained on the synthetic dataset with the representation expert. On the contrary, the classifier expert $\mathcal{M}_{\mathcal{B}}$ only trains the classifier layer, so we also only train the student classifier layer and only match the classifier layer during the classifier matching. During the whole matching process, we jointly match the representation trajectory and the classifier trajectory using the following loss:
\begin{align}
\label{loss:overall}
    \mathcal{L} = & \, \lambda_{\text{rep}}\mathcal{L}_{\text{match}}(\hat{\theta}_{t+N},\theta^{\mathcal{D}*}_{t+M},\theta^{\mathcal{D}*}_{t}) \nonumber \\
    & + \lambda_{\text{cls}}\mathcal{L}_{\text{match}}(\hat{\theta}_{t+N},\theta^{\mathcal{B}*}_{t+M},\theta^{\mathcal{B}*}_{t}),
\end{align}
where $\lambda_{\text{rep}}$ and $\lambda_{\text{cls}}$ are hyper-parameters that control the weighting ratio, $\hat{\theta}_{t+N}$ is the student model at its training epoch $t+N$, $\theta^{\mathcal{D}*}_{t}$ and $\theta^{\mathcal{B}*}_{t}$ are the representation backbone and classifier of the expert at epoch $t$, respectively. 
$\hat{\theta}_{t+N}$ is updated by our proposed long-tailed distributed loss $\mathcal{L}^\text{c}$ in Equation \ref{eq:eq3}, and $\mathcal{L}_{\text{match}}$ follows the form in Equation \ref{eq:tr_loss}.

%% file: sec/3_experiments.tex
\section{Experiments}
\label{sec:experiments}

\subsection{Experiment Datasets}
We evaluate our methods on four widely used long-tailed datasets of different scales: CIFAR-10-LT, CIFAR-100-LT \cite{cui2019classcifarlt}, TinyImageNet-LT, and ImageNet-LT~\cite{liu2019largeimagenetlt}. The experiments are conducted under various degrees of imbalance and IPC settings. 
Following~\cite{du2023globalGLMC,cui2019classcifarlt, liu2019largeimagenetlt}, these long-tailed datasets are sampled from the original balanced CIFAR~\cite{cifar} and ImageNet~\cite{deng2009imagenet} datasets. The sampling procedure is done by randomly selecting different amounts of images from each class, where the image number $|\mathcal{D}_c|$ for class $c$ is determined by an exponential decay function $|\hat{\mathcal{D}}_c|=|\mathcal{D}_c|\mu^{c}$, and $\mu^{c}=\beta^{-(c/C)}$, where $C$ is the number of classes and $\beta$ is the imbalanced factor. The degree of imbalance is measured via the imbalanced factor $\beta = \mathcal{D}_{0}/\mathcal{D}_{C}$~\cite{cui2019classcifarlt}. 
The larger $\beta$ is, the more imbalanced the dataset is.

\subsection{Main Results}

\begin{table*}
\centering
\resizebox{1\linewidth}{!}{
\begin{tabular}{l|ccc|ccc|ccc|ccc}
\toprule
Dataset & \multicolumn{6}{c|}{CIFAR-100-LT} & \multicolumn{6}{c}{Tiny-ImageNet-LT} \\
\midrule
Imbalance Factor & \multicolumn{3}{c|}{10} & \multicolumn{3}{c|}{20} & \multicolumn{3}{c|}{10} & \multicolumn{3}{c}{20}\\
IPC & 10 & 20 & 50 & 10 & 20 & 50 & 10 & 20 & 50 & 10 & 20 & 50 \\
\midrule
Random & 14.2$\pm$0.6 & 21.7$\pm$0.6 & 32.1$\pm$0.6 & 15.0$\pm$0.3 & 21.6$\pm$0.5 & 30.5$\pm$0.5 & 7.4$\pm$0.2 & 13.5$\pm$0.4 & 20.6$\pm$0.4 & 7.6$\pm$0.1 & 13.2$\pm$0.4 & 19.8$\pm$0.1\\
K-Center Greedy~\cite{sener2017activekcenter} & 10.7$\pm$0.9 & 15.9$\pm$1.0 & 24.8$\pm$0.2 & 10.0$\pm$0.5 & 15.1$\pm$0.6 & 23.8$\pm$0.3 & 10.4$\pm$1.5 & 11.3$\pm$1.2 & 7.7$\pm$1.0 & 12.6$\pm$1.4 & 9.9$\pm$2.2 & 13.2$\pm$2.7\\
Graph-Cut~\cite{iyer2021submodulargraphcut} & 16.9$\pm$0.3 & 22.2$\pm$0.4 & 29.9$\pm$0.4 & 16.0$\pm$0.5 & 20.7$\pm$0.5 & 28.7$\pm$0.3 & 9.8$\pm$0.7 & 5.6$\pm$0.6 & 10.9$\pm$1.1 & 3.4$\pm$0.8 & 5.0$\pm$1.0 & 4.2$\pm$1.1\\
\midrule
DC~\cite{zhao2020datasetdc} & 24.0$\pm$0.3 & 27.4$\pm$0.3 & 27.4$\pm$0.3 & 23.2$\pm$0.3 & 26.2$\pm$0.3 & 27.4$\pm$0.3 & - & - & - & - & - & -\\
MTT~\cite{cazenavette2022datasetmtt} & 14.3$\pm$0.1 & 16.7$\pm$0.2 & 13.8$\pm$0.2 & 12.6$\pm$0.3 & 15.0$\pm$0.2 & 10.6$\pm$0.5 & 11.1$\pm$0.2 & 18.1$\pm$0.2 & 23.1$\pm$0.1 & 7.7$\pm$0.1 & 14.7$\pm$0.2 & 15.6$\pm$0.3\\
DREAM~\cite{dream} & 10.1$\pm$0.4 & 12.0$\pm$1.0 & 13.1$\pm$0.4 & 9.4$\pm$0.4 & 10.3$\pm$0.6 & 12.3$\pm$0.3 & 5.4$\pm$0.3 & 6.8$\pm$0.1 & 7.8$\pm$0.2 & 4.8$\pm$0.1 & 6.0$\pm$0.2 & 7.4$\pm$0.1 \\
DATM~\cite{datm} & 28.2$\pm$0.4 & 34.1$\pm$0.2 & 31.6$\pm$0.1 & 25.3$\pm$0.3 & 27.2$\pm$0.1 & 27.1$\pm$0.1 & 21.3$\pm$0.1 & 14.5$\pm$0.1 & 26.8$\pm$0.1 & 14.0$\pm$0.6 & 19.0$\pm$0.3 & 23.1$\pm$0.1 \\
\cellcolor{lightblue}\textbf{Ours} & \cellcolor{lightblue}\textbf{31.5$\pm$0.2} & \cellcolor{lightblue}\textbf{37.5$\pm$0.4} & \cellcolor{lightblue}\textbf{40.0$\pm$0.1} & \cellcolor{lightblue}\textbf{31.4$\pm$0.5} & \cellcolor{lightblue}\textbf{35.1$\pm$0.4} & \cellcolor{lightblue}\textbf{37.0$\pm$0.7} & \cellcolor{lightblue}\textbf{26.0$\pm$0.3} & \cellcolor{lightblue}\textbf{27.9$\pm$0.2} & \cellcolor{lightblue}\textbf{30.3$\pm$0.2} & \cellcolor{lightblue}\textbf{23.6$\pm$0.3} & \cellcolor{lightblue}\textbf{25.5$\pm$0.3} & \cellcolor{lightblue}\textbf{28.0$\pm$0.6}\\
\midrule
Full Dataset & \multicolumn{3}{c|}{40.3$\pm$0.4} & \multicolumn{3}{c|}{36.5$\pm$0.4} & \multicolumn{3}{c|}{31.2$\pm$1.0} & \multicolumn{3}{c}{28.3$\pm$0.3} \\
\bottomrule[1pt]
\end{tabular}
}
\vspace{-0.3cm}
\caption{\textbf{Quantitative comparisons with the SOTA methods on CIFAR-100-LT and Tiny-ImageNet-LT.} When scaling up to larger datasets, our approach still outperforms all the included baselines, and achieves nearly lossless performance under several settings.
}
\label{tab:cifar100}
  \vspace{-14pt}
\end{table*}

\begin{table}
\centering
\resizebox{1\linewidth}{!}{
\begin{tabular}{l|cc|cc}
\toprule
Dataset & \multicolumn{4}{c}{ImageNet-LT} \\
\midrule
Imbalance Factor & \multicolumn{2}{c|}{5} & \multicolumn{2}{c}{10} \\
IPC & 10 & 20 & 10 & 20 \\
\midrule
Random & 3.9$\pm$0.1 & 7.0$\pm$0.1 & 3.9$\pm$0.1 & 6.8$\pm$0.1 \\
G-VBSM~\cite{shao2024generalized} & - & 1.0$\pm$0.1 & - & 1.0$\pm$0.1 \\
TESLA \cite{cui2023scalingtesla} & 3.0$\pm$0.1 & - & 2.7$\pm$0.1 & -  \\
DATM~\cite{datm} & 7.4$\pm$0.1 & 8.1$\pm$0.2 & 7.9$\pm$0.1 & 8.2$\pm$0.1  \\
SRe$^2$L~\cite{yin2023squeeze} & 6.7$\pm$0.1 & 10.1$\pm$0.1 & 7.7$\pm$0.3 & 10.9$\pm$1.0 \\
\rowcolor{lightblue}\textbf{Ours} & \textbf{20.8$\pm$0.2} & \textbf{21.0$\pm$0.1} & \textbf{20.3$\pm$0.1} & \textbf{20.7$\pm$0.1} \\
\midrule
Full Dataset & \multicolumn{2}{c|}{27.6$\pm$0.3} & \multicolumn{2}{c}{26.6$\pm$0.2}\\
\bottomrule[1pt]
\end{tabular}
}
\vspace{-0.3cm}
\caption{\textbf{Quantitative comparisons on ImageNet-LT.} The missing results are due to out-of-memory. 
}
\label{tab:imagenet}
  \vspace{-17pt}
\end{table}

\textbf{CIFAR-10-LT.} 
The experiment results on CIFAR-10-LT are presented in Table~\ref{tab:cifar10}, showing that our proposed method consistently outperforms all other dataset distillation baselines. 
Table~\ref{tab:cifar10} further highlights the advantages of our approach on long-tailed datasets.
When the imbalance factor is low, prior dataset distillation methods like DATM perform reasonably well. For instance, DATM only experiences a 10.0\% drop in accuracy when $\beta=10$ and IPC=50, compared to the accuracy achieved with the full dataset. In these scenarios, our method surpasses the best baseline, DATM, by 0.9\% on $\beta=10$ and 3.8\% on $\beta=50$. 

When the imbalance factor is high (e.g., $\beta=200$), IDM~\cite{zhao2023improvedidm} achieves performance comparable to random selection. Other methods, such as MTT~\cite{cazenavette2022datasetmtt}, DREAM~\cite{dream}, DATM~\cite{datm}, and Minimax~\cite{gu2024efficientminimax}, struggle with higher imbalance factors, performing worse than random selection. In contrast, our method sustains strong performance on highly imbalanced datasets, surpassing the best existing method IDM by a substantial margin of  10.6\% on $\beta=200$. Remarkably, we achieve lossless performance on $\beta=200$ using only 50 images per class.

\noindent\textbf{CIFAR-100-LT.} The results are presented in Table~\ref{tab:cifar100}. Our method consistently outperforms all existing dataset distillation methods across all imbalance factors on CIFAR-100-LT. Similar to CIFAR-10-LT, our approach demonstrates greater improvements as the imbalance factor increases. 

\noindent\textbf{Tiny-ImageNet-LT.}
We conducted experiments on Tiny-ImageNet-LT to evaluate the scalability of our method in comparison with other baselines. The results, shown in Table~\ref{tab:cifar100}, align with our findings on CIFAR-10-LT and CIFAR-100-LT, where our method surpasses baseline methods by a growing margin as the imbalance factor increases.

\noindent\textbf{ImageNet-LT.}
We further scale up to ImageNet-LT, with results presented in Table~\ref{tab:imagenet}, where our model consistently outperforms the baselines.
Note that as the dataset size increases, many dataset distillation methods become resource-intensive and are impractical to apply.

\noindent\textbf{Cross-architecture performance.} 
Table \ref{tab:cross} presents our cross-architecture experiments, demonstrating the generalizability of our synthetic dataset. We evaluated various methods on the CIFAR-10-LT dataset with an imbalance factor of $\beta=100$ and IPC=50, comparing against random selection, MTT, and DATM. As shown in the table, our distilled dataset consistently outperforms other methods by a significant margin. Specifically, our method achieves a 12.7\% accuracy improvement on ConvNet, a 17.1\% improvement on ResNet-18, a 7.7\% improvement on VGG-11, and a 6\% improvement on AlexNet. The results also indicate that the imbalance distilled by ConvNet in traditional DD methods negatively impacts the performance of other models. Similar to experiments on ConvNets, MTT and DATM perform poorly on unseen models due to the imbalanced dataset and the biased expert trained on it, often yielding worse results than random selection.

\begin{table}[t]
    \centering
    \resizebox{1\linewidth}{!}{
        \begin{tabular}{l|cccc}
        \toprule
        Method & ConvNet-3 & ResNet-18 & VGG-11 & AlexNet \\ \midrule
        Random & 52.0 & 49.2 & 47.8 & 47.2 \\
        MTT \cite{cazenavette2022datasetmtt} & 49.1 & 42.9 & 42.3 & 40.0 \\
        DATM \cite{datm} & 44.4 & 42.4 & 42.2 & 44.9 \\
        \rowcolor{lightblue}\textbf{Ours} & \textbf{64.7} & \textbf{60.0} & \textbf{55.5} & \textbf{53.2} \\
        \bottomrule
        \end{tabular}
        }
    \vspace{-0.3cm}
    \caption{\textbf{Cross-architecture evaluation on CIFAR-10-LT.}}
    \vspace{-16pt}
    \label{tab:cross}
\end{table}

\begin{figure*}[t]
  \centering
  \begin{subfigure}{0.19\linewidth}
    \centering
    \captionsetup{justification=centering}
    \includegraphics[trim=0 0 0 100,clip,width=\linewidth]{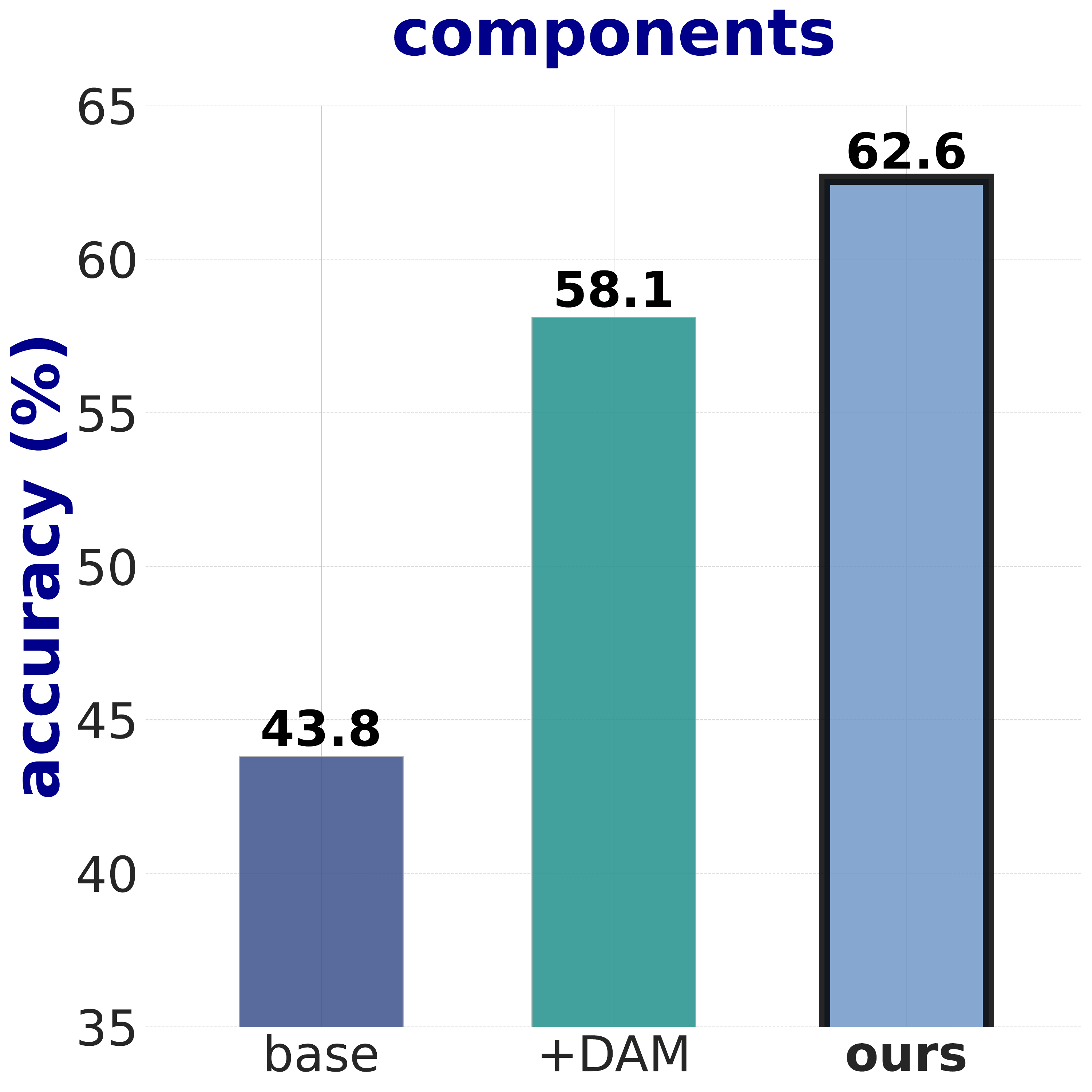}
    \vspace{-0.2in}
    \caption{Effect of the different components}
    \label{fig:abl_components}
  \end{subfigure}
  \hfill
  \centering
  \begin{subfigure}{0.19\linewidth}
    \centering
    \captionsetup{justification=centering}
    \includegraphics[trim=0 0 0 100,clip,width=\linewidth]{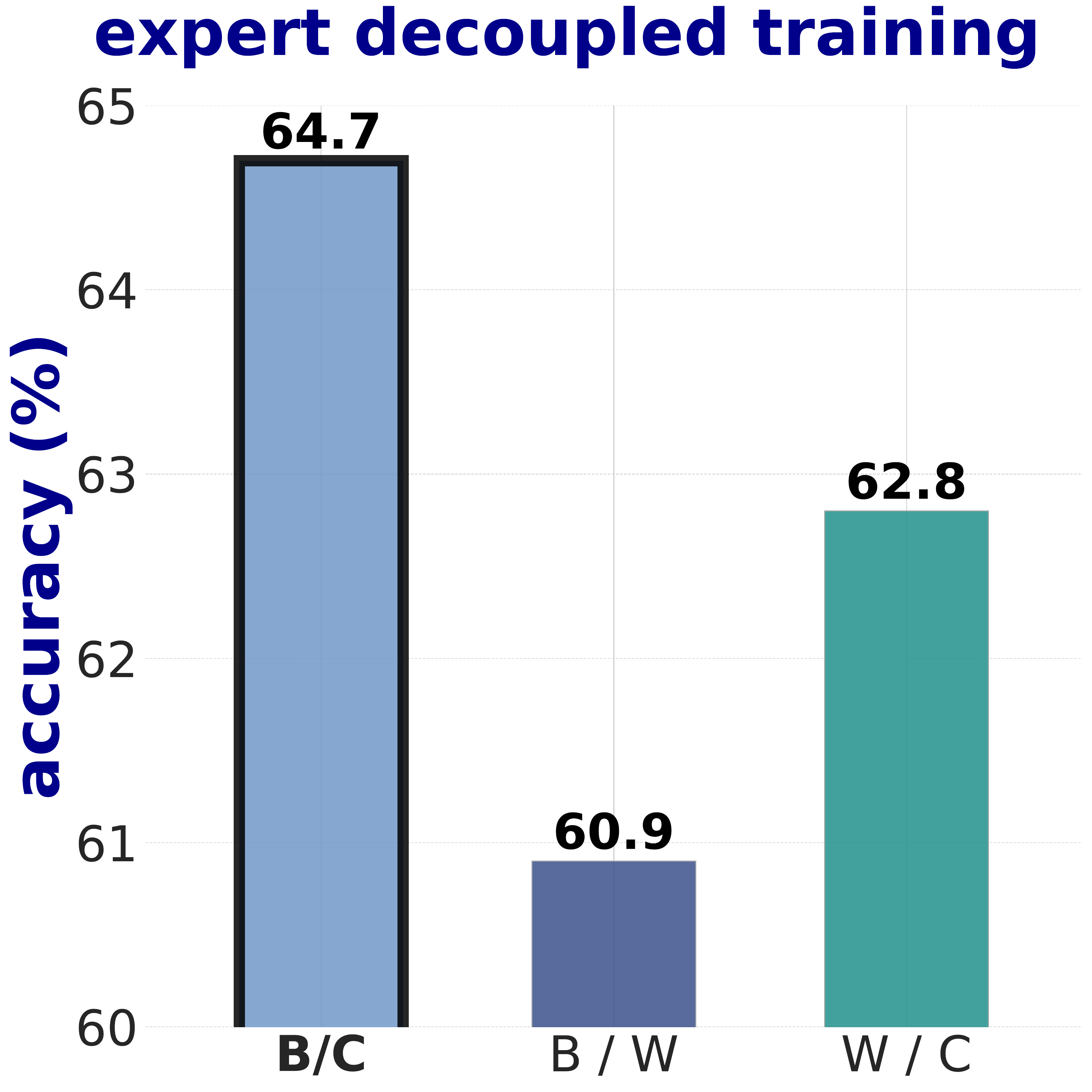}
    \vspace{-0.2in}
    \caption{Comparison of different matching strategies}
    \label{fig:abl_expert}
  \end{subfigure}
  \hfill
  \centering
  \begin{subfigure}{0.19\linewidth}
    \centering
    \captionsetup{justification=centering}
    \includegraphics[trim=0 0 0 100,clip,width=\linewidth]{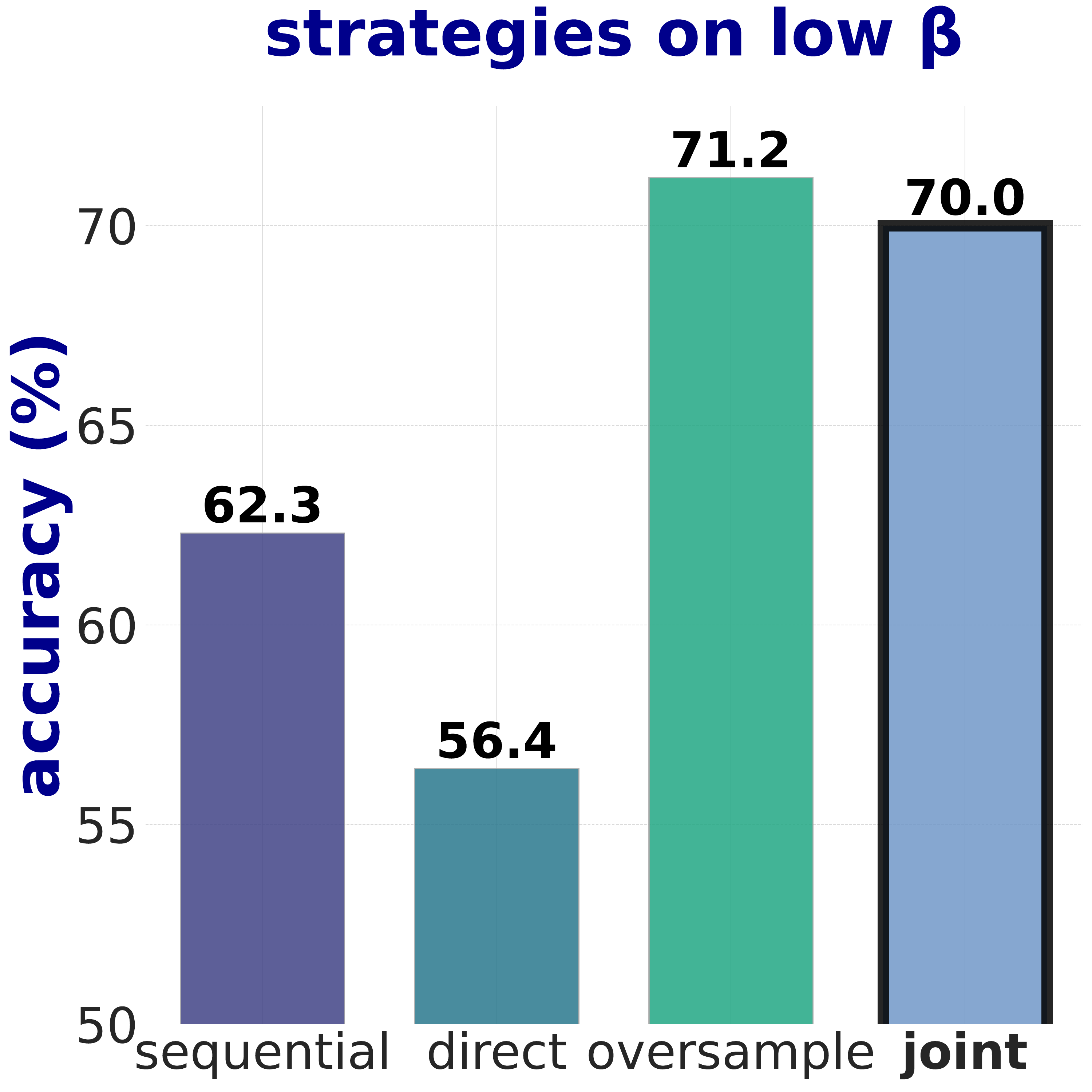}
    \vspace{-0.2in}
    \caption{Comparison of matching strategies on low $\beta$}
    \label{fig:abl_matching_1}
  \end{subfigure}
  \hfill
  \centering
  \begin{subfigure}{0.19\linewidth}
    \centering
    \captionsetup{justification=centering}
    \includegraphics[trim=0 0 0 100,clip,width=\linewidth]{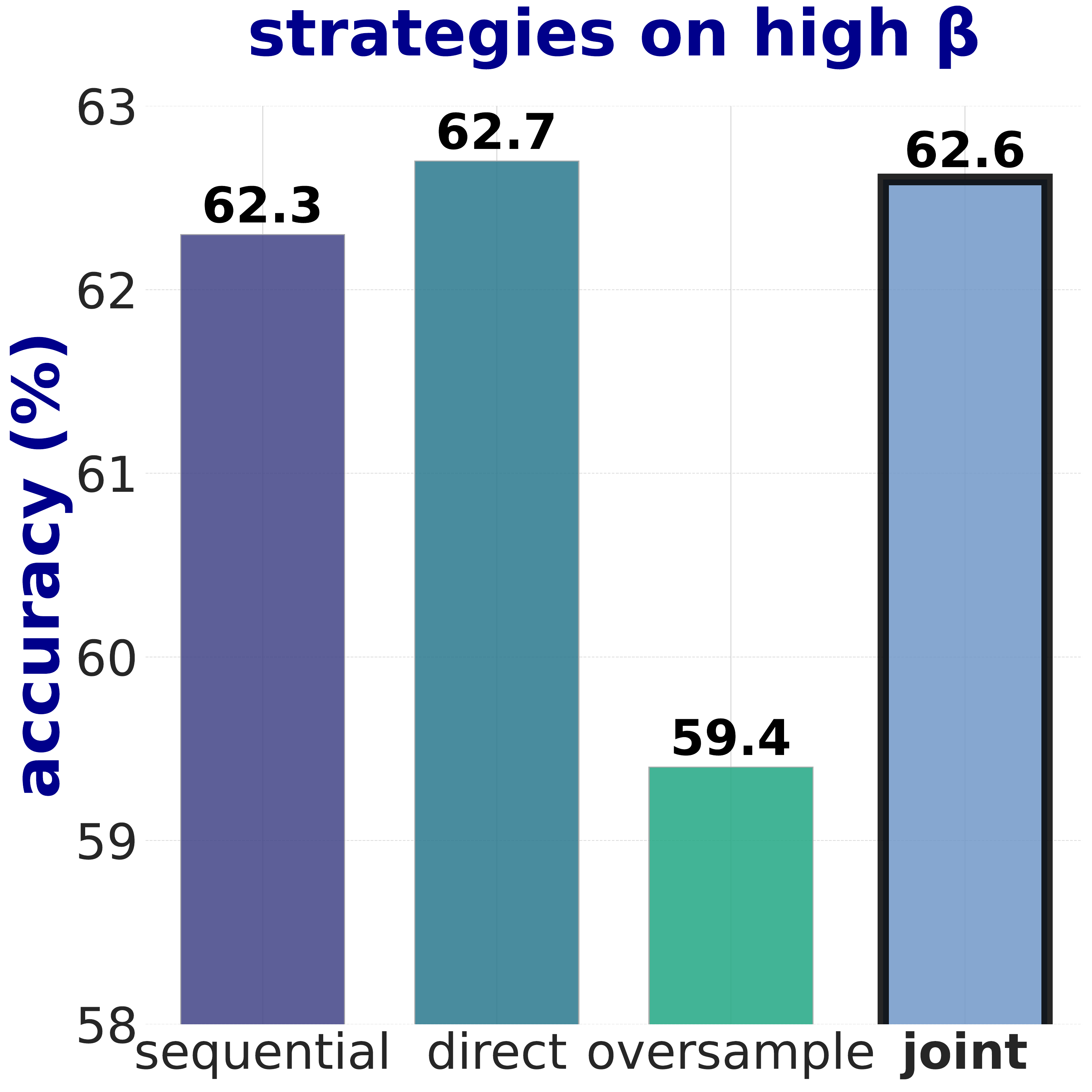}
    \vspace{-0.2in}
    \caption{Comparison of matching strategies on high $\beta$}
    \label{fig:abl_matching_2}
  \end{subfigure}
  \hfill
  \centering
  \begin{subfigure}{0.19\linewidth}
    \centering
    \captionsetup{justification=centering}
    \includegraphics[trim=0 0 0 100,clip,width=\linewidth]{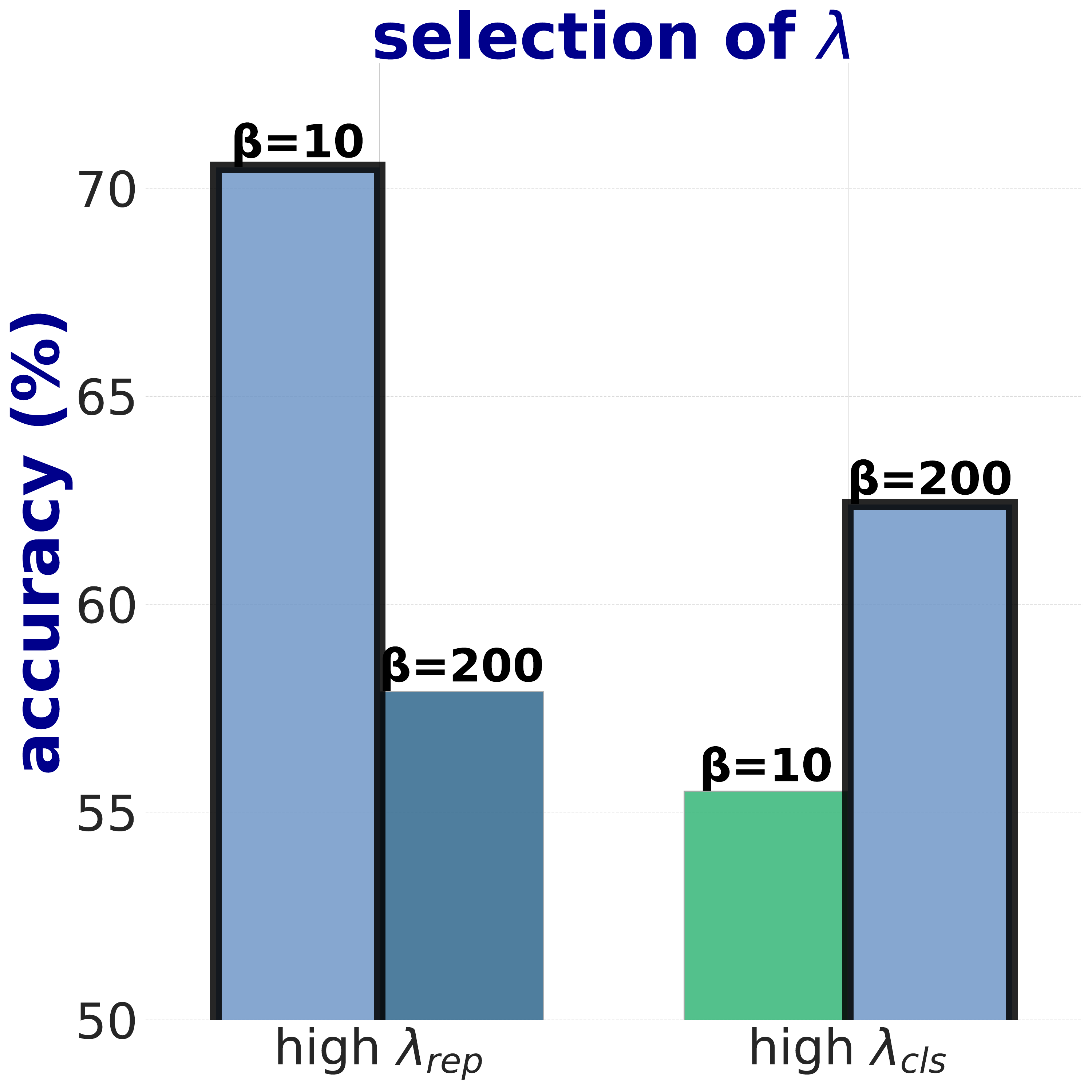}
    \vspace{-0.2in}
    \caption{Effect of the relative scale of $\lambda_{\text{rep}}$ and $\lambda_{\text{cls}}$}
    \label{fig:abl_selection}
  \end{subfigure}
  \hfill
  \vspace{-0.1in}
  \caption{(a) Ablation study on the proposed components, Distribution-agnostic Matching and Expert Decoupling. (b) Ablation of different matching strategies with expert model layers. (c) \& (d) Comparisons of matching strategies for long-tailed dataset distillation under low imbalance factor (c) and high imbalance factor (d). (e) Ablation study on hyperparameters $\lambda_{\text{rep}}$ and $\lambda_{\text{cls}}$.}
\end{figure*}

\vspace{-0.1cm}

\subsection{Ablation Studies}
\label{subsec: abl}
\textbf{Ablation on different components.}
Figure~\ref{fig:abl_components} demonstrates the effectiveness of the two proposed components of our method. These experiments were performed on the CIFAR-10-LT dataset with an imbalance factor of $\beta=200$ and IPC=50. The ``base'' method is a trajectory matching method without Distribution-agnostic Matching and Expert Decoupling. ``+DAM'' refers to our method without Expert Decoupling, while ``ours'' represents the complete proposed method, with both DAM and ED applied.

The baseline, which suffers from weight inconsistency and poor expert performance as mentioned in Section~\ref{sec:intro}, only achieves 43.8\% accuracy. After incorporating Distribution-agnostic Matching, the inconsistency between student and expert weights is mitigated, resulting in a 14.3\% performance improvement. Further integrating Expert Decoupling boosts the performance to 62.6\%.

\noindent\textbf{Effect of classifier layer matching.} 
As discussed in Section~\ref{sec:ed}, 
during representation learning, the model is influenced by the long-tailed distribution, resulting in a biased classifier. During classifier fine-tuning, the classifier is fine-tuned on balanced data with the backbone frozen. Therefore, it is crucial to explore strategies for matching each expert's different modules during dataset distillation.
The experiments are presented in Figure~\ref{fig:abl_expert}. ``B", ``C", and ``W" refer to matching with the representation backbone, the classifier layer, and the whole model, respectively. ``B/C" refers to matching the student backbone with the representation expert's backbone, and the student classifier layer with the classification expert's classifier. 

As shown in Figure~\ref{fig:abl_expert}, the best performance is achieved by matching the backbone with representation experts and the classifier layer with classification experts. This outcome aligns with the principles of decoupled learning. Since the classifier layer of representation experts is heavily affected by the long-tailed dataset, including it in the matching process causes a performance drop. In addition, matching the entire classifier expert negatively impacts performance, as only the classifier layer is trained at this stage.

\noindent\textbf{Effect of different matching strategies.} 
When distilling a long-tailed dataset, various intuitive strategies can be employed with decoupling trained experts. We explored several strategies, including sequential matching, classification expert matching, oversampled dataset, and joint matching.
Sequential matching first matches representation experts, followed by classification experts. Classification expert matching only matches with the classification expert. The oversampled dataset approach involves oversampling the tail classes before distilling the dataset rather than using the decoupling strategy. Joint matching, used in our Expert Decoupling, involves matching both experts simultaneously in the outer loop of the matching.

As shown in Figures \ref{fig:abl_matching_1} and \ref{fig:abl_matching_2}, joint matching (joint), which we use in Expert Decoupling, achieves the best overall performance. Sequential matching (sequential) and directly matching with classification experts (direct) perform poorly when the imbalance factor is low, meaning the data is ``not very imbalanced." Sequential matching suffers due to the trajectory gap between the two types of experts. Classification expert matching is less effective when the imbalance factor is low because the representation layers learn more information in such scenarios, and solely matching the classification layers fails to capture sufficient information.
As depicted in Figure \ref{fig:abl_matching_2}, the oversampled dataset strategy (oversample) performs poorly when the imbalance factor is high. This finding aligns with previous work \cite{cui2019classcifarlt}, which indicates that oversampling leads to model overfitting.

\begin{figure*}[t]
  \centering
  \begin{subfigure}{0.24\linewidth}
    \centering
    \captionsetup{justification=centering}
    \includegraphics[trim=0 0 0 100,clip,width=\linewidth]{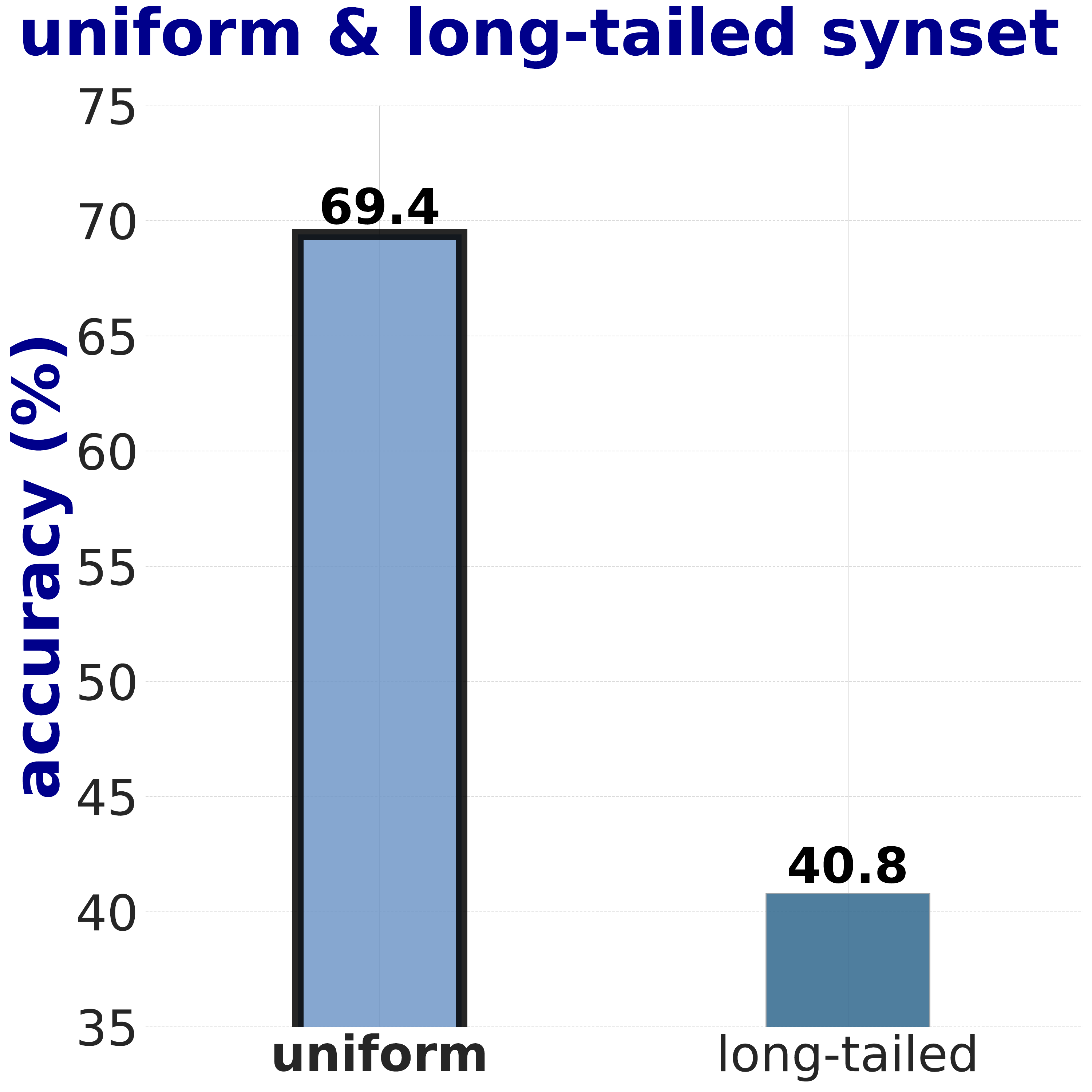}
    \vspace{-0.2in}
    \caption{Effect of data distribution}
    \label{fig:anl_datadistribution}
  \end{subfigure}
  \hfill
  \centering
  \begin{subfigure}{0.24\linewidth}
    \centering
    \captionsetup{justification=centering}
    \includegraphics[trim=0 0 0 100,clip,width=\linewidth]{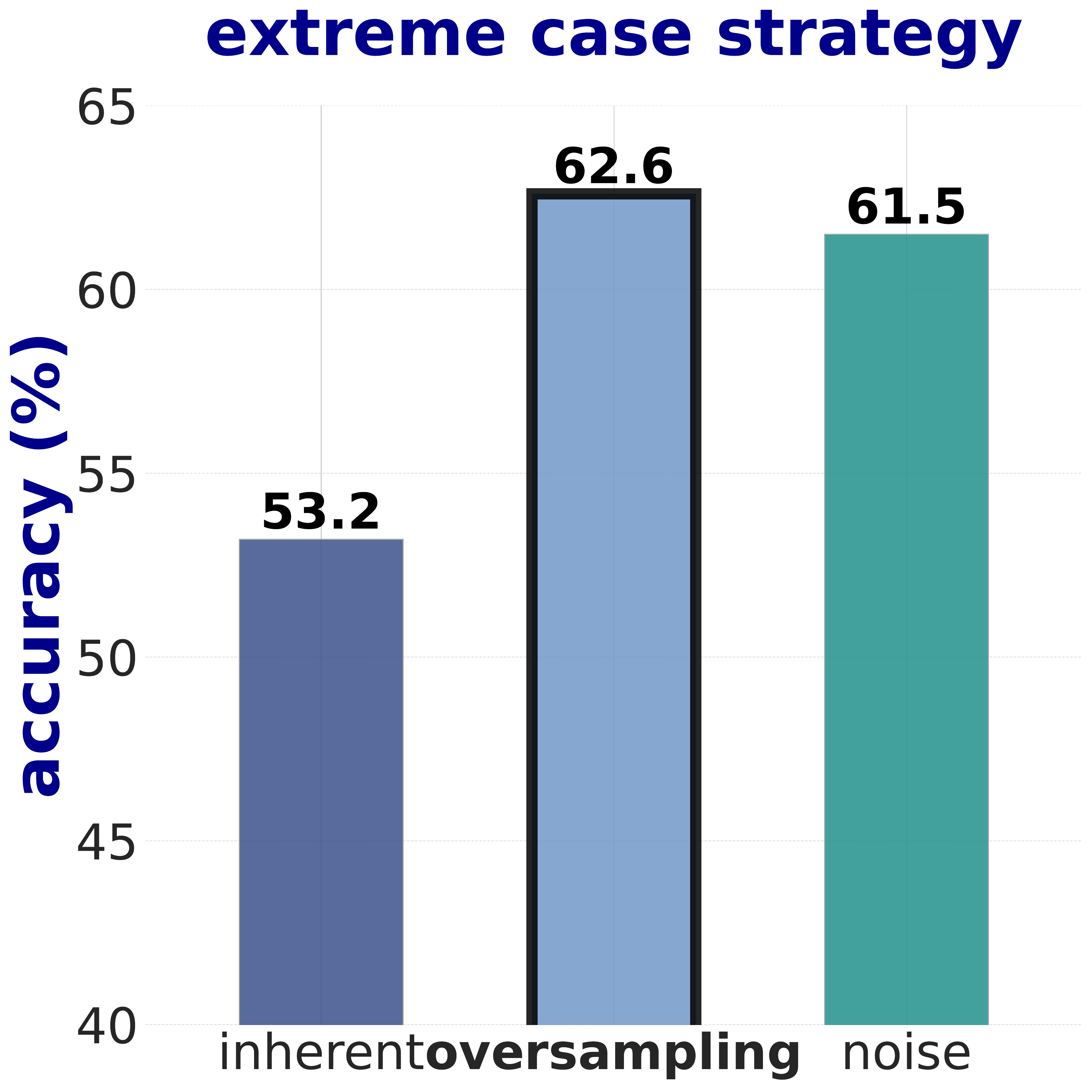}
    \vspace{-0.2in}
    \caption{Strategies for extreme cases}
    \label{fig:anl_extreme}
  \end{subfigure}
  \hfill
  \centering
  \begin{subfigure}{0.24\linewidth}
    \centering
    \captionsetup{justification=centering}
    \includegraphics[trim=0 0 0 100,clip,width=\linewidth]{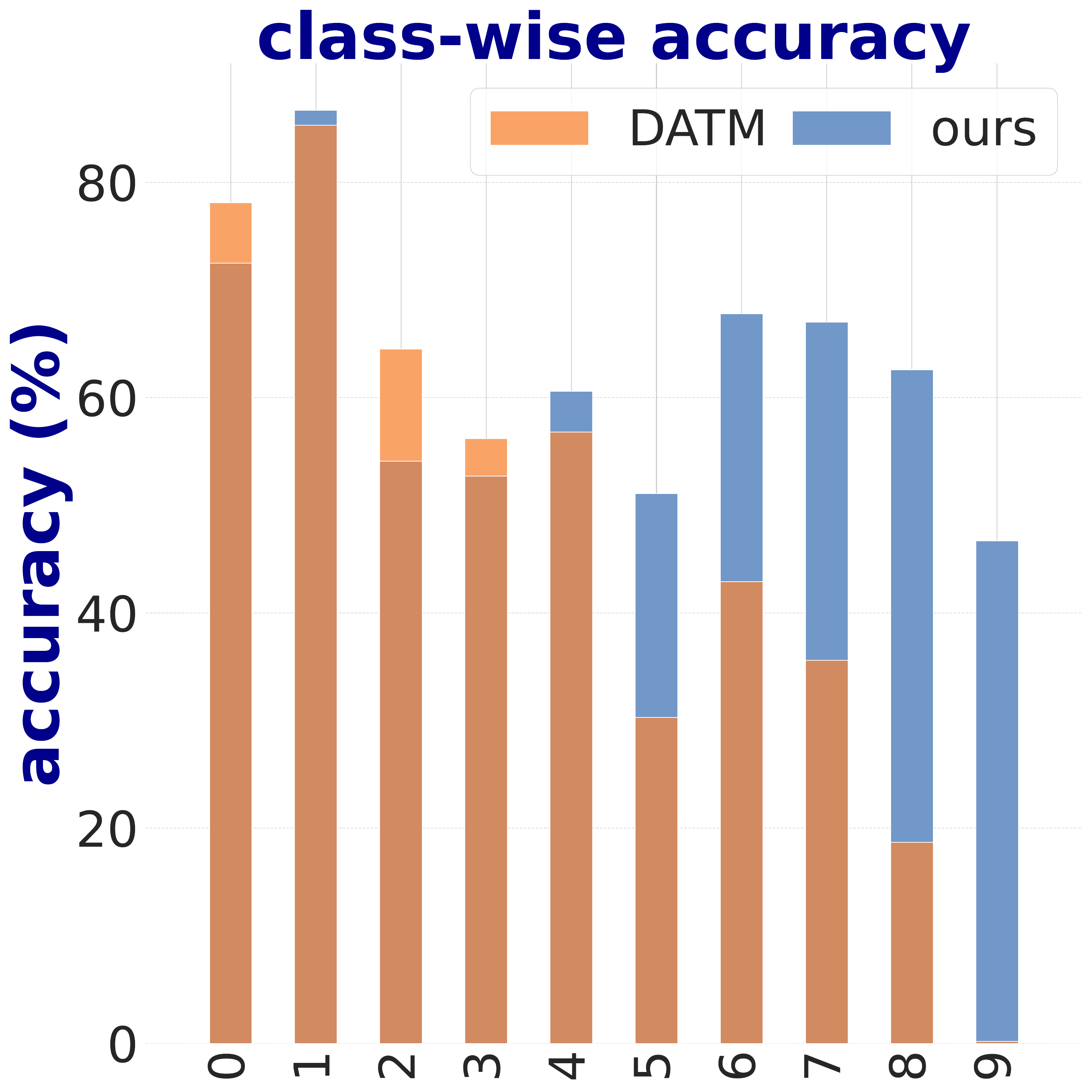}
    \vspace{-0.2in}
    \caption{Comparison with DATM}
    \label{fig:anl_class}
  \end{subfigure}
  \hfill
  \centering
  \begin{subfigure}{0.24\linewidth}
    \centering
    \captionsetup{justification=centering}
    \includegraphics[trim=0 0 0 100,clip,width=\linewidth]{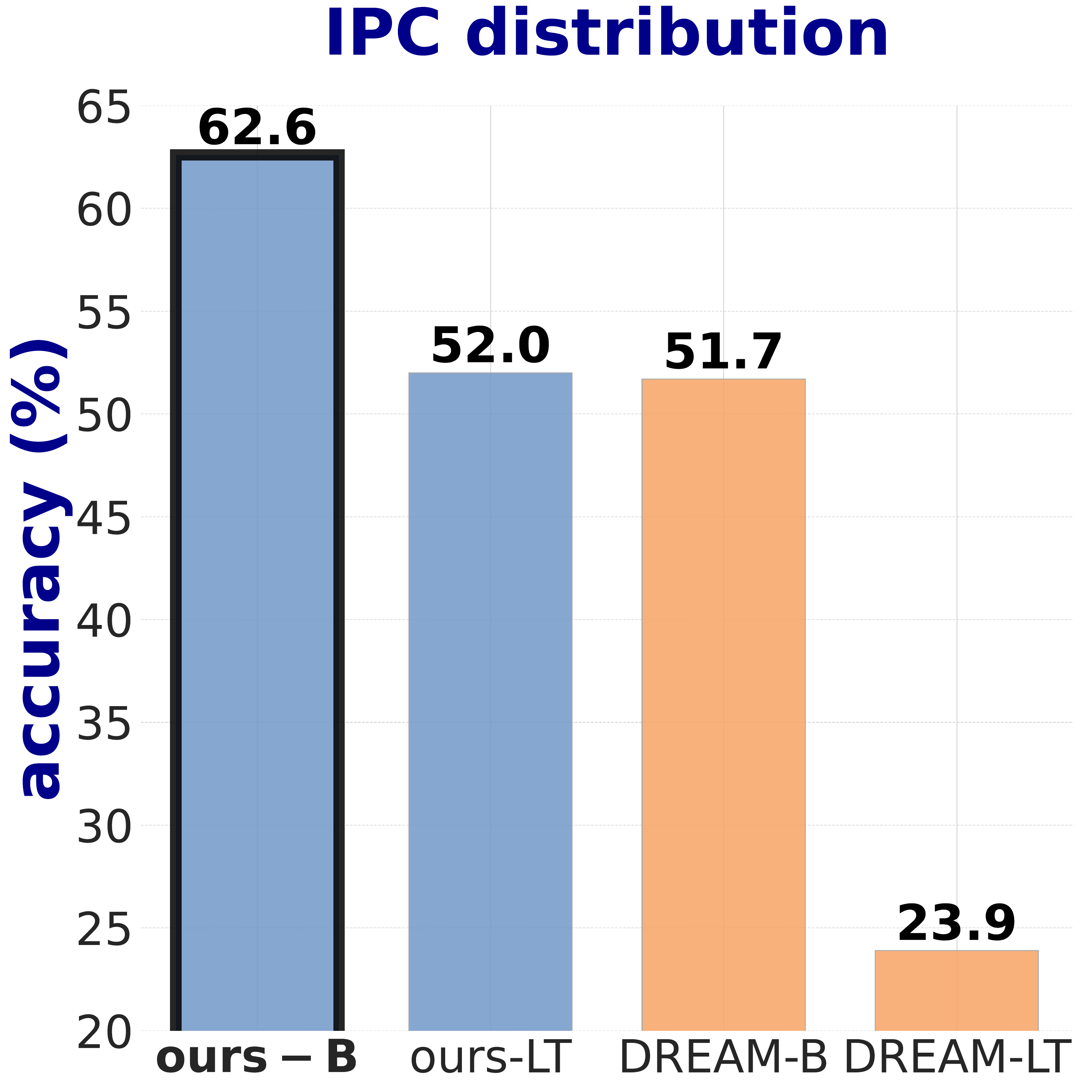}
    \vspace{-0.2in}
    \caption{Effect of distilled distribution}
    \label{fig:anl_ipc}
  \end{subfigure}
  \hfill
  \vspace{-5pt}
  \caption{\textbf{(a)} Comparison of DD performance between a uniform dataset and a long-tailed dataset with the same number of samples. \textbf{(b)} Comparison of strategies to handle an extremely low number of samples. \textbf{(c)} Comparison of class-wise accuracy between DATM and our method. \textbf{(d)} Comparison between balanced IPC and long-tailed distributed IPC.}
  \vspace{-0.5cm}
\end{figure*}

\noindent\textbf{Effect of hyperparameters.} 
The most important hyperparameters in our method are $\lambda_{\text{rep}}$ and $\lambda_{\text{cls}}$ in Equation \ref{loss:overall}.
Together, they regulate the balance between the matching representation expert and the classification expert.
The setting of $\lambda_{\text{rep}}$ and $\lambda_{\text{cls}}$ depends on the imbalance factors.
From Figure~\ref{fig:abl_selection}, we have the following observations. For a ``more imbalanced'' dataset (e.g., $\beta=200$), lower $\lambda_{\text{rep}}$ and higher $\lambda_{\text{cls}}$ are preferred. For ``less imbalanced'' datasets (e.g., $\beta=10$), higher $\lambda_{\text{rep}}$ and lower $\lambda_{\text{cls}}$ are chosen. This observation is consistent with our experiments in Figure~\ref{fig:abl_matching_1} and \ref{fig:abl_matching_2}, where only matching classifier experts perform well when $\beta$ is high. This is because the representation layers learn more information when $\beta$ is low, and increasing $\lambda_{\text{rep}}$ leverages information from the representation layers.

\begin{table}[t]
    \centering
    \begin{minipage}{.48\linewidth} 
      \centering
      \resizebox{\linewidth}{!}{
      \begin{tabular}{c|c}
        \toprule
        Hyperparameters & $\beta=10$ \\
        \midrule
        $\lambda_{\text{rep}}=1, \lambda_{\text{cls}}=0.2$ & 70.5 \\
        $\lambda_{\text{rep}}=1, \lambda_{\text{cls}}=0.4$ & 68.9 \\
        $\lambda_{\text{rep}}=1, \lambda_{\text{cls}}=0.6$ & 68.0 \\
        \bottomrule
      \end{tabular}}\
      \vspace{-0.2cm}
    \end{minipage}
    \begin{minipage}{.49\linewidth} 
      \centering
      \resizebox{\linewidth}{!}{
      \begin{tabular}{c|c}
        \toprule
        Hyperparameters & $\beta=200$ \\
        \midrule
        $\lambda_{\text{rep}}=0.2, \lambda_{\text{cls}}=1$ & 62.4 \\
        $\lambda_{\text{rep}}=0.4, \lambda_{\text{cls}}=1$ & 61.7 \\
        $\lambda_{\text{rep}}=0.6, \lambda_{\text{cls}}=1$ & 61.3 \\
        \bottomrule
      \end{tabular}}\
      \vspace{-0.2cm}
    \end{minipage}
    \vspace{-0.1cm}
    \caption{Effect of $\lambda_{\text{rep}}$ and $\lambda_{\text{cls}}$ under different imbalance factors.}
    \label{tab:lambda}
    \vspace{-0.4cm}
\end{table}

We further analyze the sensitivity of the hyperparameters in Table~\ref{tab:lambda}. We can observe a minor change in the performance as we change hyperparameters rapidly. When the relative magnitude of the two hyperparameters is preserved, their absolute values do not affect the distillation performance much, suggesting the robustness of our method.

\subsection{Long-tailed Dataset Distillation Analysis}
\noindent\textbf{Effect of imbalance in long-tailed dataset distillation.}
In this part, we empirically show that the performance drop in long-tailed dataset distillation is due to the data imbalance instead of the total sample number reduction of the target dataset. We compare the performance of using regular dataset distillation methods for CIFAR-10-LT (with $\beta=200$ and IPC=50) and a subset of CIFAR-10 that contains the same amount of data but in a uniform distribution. The total number of samples is 11203 under this setting. 

As we can observe in Figure~\ref{fig:anl_datadistribution}, the performance gap between distilling a long-tailed dataset and a balanced subset can be as large as 28.6\%, even though the two target datasets have the same total amount of samples. Thus, we conclude that it is the imbalanced distribution in the target dataset that prevents effective distillation. Based on this observation, we delved into it and proposed methods to alleviate the imbalanced distribution's negative effects.

\noindent\textbf{Handling classes with extremely small number of samples.}
In long-tailed dataset distillation, a scenario exists where the number of samples in tail classes is even less than the required number of images per class. In this case, we employ and compare three strategies: 1) Inherent strategy, which keeps the number of samples the same in the synthetic dataset; 2) Oversampling, which duplicates the samples of the tail classes before distillation; 3) Noise generation, which randomly generates Gaussian noises to supplement the insufficient samples for tail classes. From figure~\ref{fig:anl_extreme}, we see that the inherent strategy provides a comparatively lower accuracy, while oversampling and noise generation achieve better and similar performance. 

\noindent\textbf{Class-wise accuracy for long-tailed dataset.}
We also provide class-wise accuracy to illustrate the impact of the imbalanced data, and we show that our method can effectively improve the accuracy of tail classes. As shown in Figure~\ref{fig:anl_class}, DATM performs poorly on tail classes, and even achieves near-zero performance on the last class. On the contrary, our method can successfully boost the tail class accuracy while reserving performance on head classes.

\noindent\textbf{Choice of distilled dataset distribution.}
Should we distill a long-tailed dataset into another long-tailed dataset or a balanced one? We compared the performance of distilling the target dataset into either a similarly distributed long-tailed synthetic dataset or a balanced synthetic dataset using our method and DREAM~\cite{dream}. For the long-tailed synthetic dataset, we maintained the same total number of samples as in the synthetic dataset with IPC=50. As shown in Figure \ref{fig:anl_ipc}, both methods demonstrate that distilling the dataset into a balanced synthetic dataset provides better performance than distilling it into a long-tailed one. This result is reasonable since the model trained on the synthetic long-tailed dataset will be affected by the long-tailed distribution.

%% file: sec/4_related_work.tex
\vspace{-0.1cm}
\section{Related Work}
\label{sec:related_work}
\subsection{Dataset Distillation}
Dataset distillation~\cite{wang2018dataset} was first formulated by Wang et al. 
The goal of dataset distillation is to synthesize a small dataset, such that models trained on it have good performance on the original dataset. The following works can be divided into kernel ridge regression-based methods~\cite{loo2022efficientkernel1,nguyen2020datasetkernel3, zhou2022datasetkernel2}, gradient matching methods~\cite{datm,dream,zhao2020datasetdc,cui2023scalingtesla,wang2024emphasizing}, representation matching methods~\cite{shang2024mim4dd,zhao2023improvedidm}, and generative-based methods~\cite{gu2024efficientminimax, su2024d4m}.
Specifically, DATM~\cite{datm} achieves lossless performance by aligning the difficulty of the generated patterns. 
Tesla~\cite{cui2023scalingtesla} significantly reduces the memory required for trajectory matching.
Despite the advantages of current methods, they only focus on uniformly distributed datasets, ignoring the prevalent usage of long-tailed datasets in practice. Therefore, we propose to address the long-tailed dataset distillation problem in this paper. 
\vspace{-0.1cm}

\subsection{Long-tailed Recognition}
\vspace{-0.1cm}
Under real-world settings, data often tends to follow a long-tailed distribution~\cite{alshammari2022longweightbalancing}. Models trained on such datasets usually exhibit good performance on head classes but low accuracy on tail classes. 
A straightforward way is to pre-process the data, using oversampling~\cite{haixiang2017learning_oversample1, janowczyk2016deep_oversample3,peng2020large_oversample2}, undersampling~\cite{buda2018systematic_under1,haixiang2017learning_oversample1}, or data augmentation~\cite{chou2020remix_aug2,gidaris2018dynamic_aug1} to make it balancedly distributed. An alternative way is weight balancing~\cite{alshammari2022longweightbalancing, focal_loss, ren2020balancedsoftmaxloss}, which re-weights the layers during training to force the parameters to update unbiasedly. Decoupling methods~\cite{kang2019decoupling}
first train the model to obtain the features of the data and then fine-tune the model on the class-balanced data. In our methods, we use the decoupling method to train the expert networks during the distillation.
\vspace{-0.1cm}

%% file: sec/5_conclusion.tex
\section{Conclusion}
\label{sec:conclusion}
\vspace{-0.1cm}
We introduce long-tailed dataset distillation, a novel yet challenging task. We find that existing DD methods fail when applied to long-tailed datasets and encounter two main problems: weight distribution mismatch between the student and the expert; and suboptimal performance of the expert on the tail classes. To mitigate these issues, we propose two strategies: Distribution-agnostic Matching and Expert Decoupling. Experimental results over four long-tailed datasets demonstrate the effectiveness of our proposed method. As the pioneering work in this research area, we are the first to effectively distill long-tailed datasets.

%% file: sec/X_suppl.tex
\clearpage
\setcounter{page}{1}
\maketitlesupplementary


\section{Training Details}
\noindent\textbf{Experiment Setup.}
We compare our method with various coreset selection methods and state-of-the-art dataset distillation methods.
Consistent with these works, we use ConvNets~\cite{krizhevsky2017imagenetconvnet} for the training and evaluation. For trajectory matching methods such as MTT, DATM, and our method, we trained and saved 100 experts with 100 epochs. During evaluation, the models are trained for 1000 epochs on the synthetic dataset. For the experts of our method, the experts are trained in a decoupled manner. For representation training, the experts are trained for 100 epochs with weight decay. For classifier fine-tuning, the experts are fine-tuned for 10 epochs with MaxNorm constraint~\cite{hinton2012improvingmaxnrom1,shalev2007pegasosmaxnorm2}. 
For TinyImageNet-LT and ImageNet-LT, we use a depth-4 ConvNet in our experiments. For experiments on ImageNet-LT, we adopt the Tesla~\cite {cui2023scalingtesla} code base to reduce memory usage.

\noindent\textbf{Hyper-parameters.} We report the hyper-parameters of our method under different settings in Table~\ref{tab:hyper}. For expert epochs, image learning rate, label learning rate, and other hyper-parameters, we follow the previous works~\cite{cazenavette2022datasetmtt, datm}.

\section{More Experiments}

\subsection{Long-tailed Test Set}
We perform experiments on CIFAR10-LT so that the train/test set follows the same long-tailed distribution, and the number of the samples for each class in the test set is (1000, 555, 308, 170, 94, 52, 29, 16, 9, 5). The results are shown in Table~\ref{tab:lt_testset}. We further compare the precision, recall and F1-score of the prediction results to obtain better insights. "-" indicates that the method cannot converge during training. Our analysis reveals the following:

\begin{table}[ht]
\vspace{-0.2cm}
\centering
\resizebox{1\linewidth}{!}{
\begin{tabular}{l|cc|cc|cc|cc}
\toprule
Metrics & \multicolumn{2}{c}{Acc.(\%)} & \multicolumn{2}{c}{Precision} & \multicolumn{2}{c}{Recall} & \multicolumn{2}{c}{F1} \\
\midrule
IPC & 10 & 50 & 10 & 50 & 10 & 50 & 10 & 50 \\
\midrule
Random & 27.3$\pm$1.4 & 54.2$\pm$1.1 & 0.21$\pm$0.01 & 0.35$\pm$0.01 & 0.28$\pm$0.01 & 0.53$\pm$0.01 & 0.17$\pm$0.01 & 0.34$\pm$0.01 \\
MTT~\cite{cazenavette2022datasetmtt} & 44.7$\pm$0.0 & 12.4$\pm$1.0 & 0.04$\pm$0.00 & 0.20$\pm$0.01 & 0.10$\pm$0.00 & 0.25$\pm$0.01 & 0.06$\pm$0.00 & 0.14$\pm$0.01 \\
DATM~\cite{datm} & - & 72.7$\pm$0.5 & - & 0.42$\pm$0.01 & - & 0.47$\pm$0.01 & - & 0.44$\pm$0.01 \\
\textbf{Ours} & \textbf{51.6$\pm$1.4} & \textbf{73.2$\pm$1.2} & \textbf{0.37$\pm$0.01} & \textbf{0.46$\pm$0.01} & \textbf{0.53$\pm$0.02} & \textbf{0.62$\pm$0.01} & \textbf{0.35$\pm$0.01} & \textbf{0.48$\pm$0.01} \\
\bottomrule[1pt]
\end{tabular}
}
\vspace{-0.2cm}
\caption{\textbf{Quantitative comparisons on long-tailed test set.} 
}
\label{tab:lt_testset}
  \vspace{-0.2cm}
\end{table}

\noindent\textbf{1. The accuracy of a long-tailed testset cannot reflect the actual model performance.} Take MTT as an example; the accuracy of IPC 10 is much higher than that of IPC 50. We find that this is because the model trained in IPC 10 predicts all samples to the first class so that its accuracy reaches 44.7\% (the percentage of the first class sample numbers). However, this definitely does not indicate the model obtained by MTT IPC 10 performs well.

\noindent\textbf{2. The effectiveness of our method is consistent with the results of the balanced test set.} Instead of only focusing on accuracy, for an imbalanced test set, we should also involve precision, recall, and F1-score, as provided in Table~\ref{tab:lt_testset}. From the table, we can observe that our methods have a small leading on precision, outperform the baselines with a large margin on recall, and have the best performance on the f1 score. The results indicate that we can preserve the head class accuracy and improve the tail class accuracy. To support this conclusion, we show the class-wise accuracy comparison under IPC 50 in Table~\ref{tab:lt_testset_cls}.

\begin{table}[ht]
\vspace{-0.2cm}
\centering
\resizebox{1\linewidth}{!}{
\begin{tabular}{l|cccccccccc}
\toprule
Method & cls0 & cls1 & cls2 & cls3 & cls4 & cls5 & cls6 & cls7 & cls8 & cls9 \\
\midrule
DATM~\cite{datm} & \textbf{78.1} & 86.3 & \textbf{58.8} & \textbf{60.0} & 57.4 & 38.5 & 51.7 & 31.3 & 22.2 & 0.0 \\
Ours & 76.8 & \textbf{91.4} & 58.1 & 53.5 & \textbf{59.6} & \textbf{53.8} & \textbf{75.9} & \textbf{68.8} & \textbf{55.6} & \textbf{20.0} \\
\bottomrule[1pt]
\end{tabular}
}
\vspace{-0.2cm}
\caption{\textbf{Class-wise accuracy on long-tailed test set.} 
}
\label{tab:lt_testset_cls}
  \vspace{-0.6cm}
\end{table}

\subsection{DD-Ranking Evaluation}
We further evaluate our method with DD-Ranking~\cite{li2024ddranking} to provide a fair evaluation for LTDD, reducing the impacts from knowledge distillation and data augmentation to reflect the real informativeness of the distilled data. We compared our method with the hard-label-based method MTT and the soft-label-based method DATM in Table~\ref{tab:ddranking}. These results indicate the effectiveness of our proposed method.

\begin{table}[ht]
\vspace{-0.2cm}
\centering
\resizebox{1\linewidth}{!}{
\begin{tabular}{l|c|c}
\toprule
Metrics & Hard Label Recovery (HLR) $\downarrow$ & Improvement Over Random $\uparrow$ \\
\midrule
MTT~\cite{cazenavette2022datasetmtt} & 31.3\% & - \\
DATM~\cite{datm} & 13.6\% & -0.3\% \\
\textbf{Ours} & \textbf{6.6}\% & \textbf{21.6}\% \\
\bottomrule[1pt]
\end{tabular}
}
\vspace{-0.2cm}
\caption{\textbf{DD-Ranking Evaluation.} 
}
\label{tab:ddranking}
  \vspace{-0.6cm}
\end{table}

\section{Compute Resources}
The computational cost comparisons of our method with the other trajectory matching methods~\cite{cazenavette2022datasetmtt, datm} are listed in Table~\ref{tab:cost}. The comparisons are done under the same hardware (NVIDIA A6000) and software environments. We can see that our computational cost is in a reasonable range.

\begin{table}[ht]
\vspace{-0.2cm}
\centering
\resizebox{1\linewidth}{!}{
\begin{tabular}{l|ccc}
\toprule
Dataset & CIFAR-10-LT & CIFAR-100-LT & TinyImageNet-LT\\
\midrule
MTT~\cite{cazenavette2022datasetmtt} & 10.0 & 40.8 & 50.2 \\
DATM~\cite{datm} & 22.7 & 98.2 & 124.0 \\
Ours & 15.0 & 56.8 & 85.9 \\
\bottomrule[1pt]
\end{tabular}
}
\vspace{-0.2cm}
\caption{\textbf{Computation cost comparison.} 
}
\label{tab:cost}
  \vspace{-0.6cm}
\end{table}

\section{Dataset Images}
\begin{figure*}[ht]
    \centering
    \includegraphics[width=\linewidth]{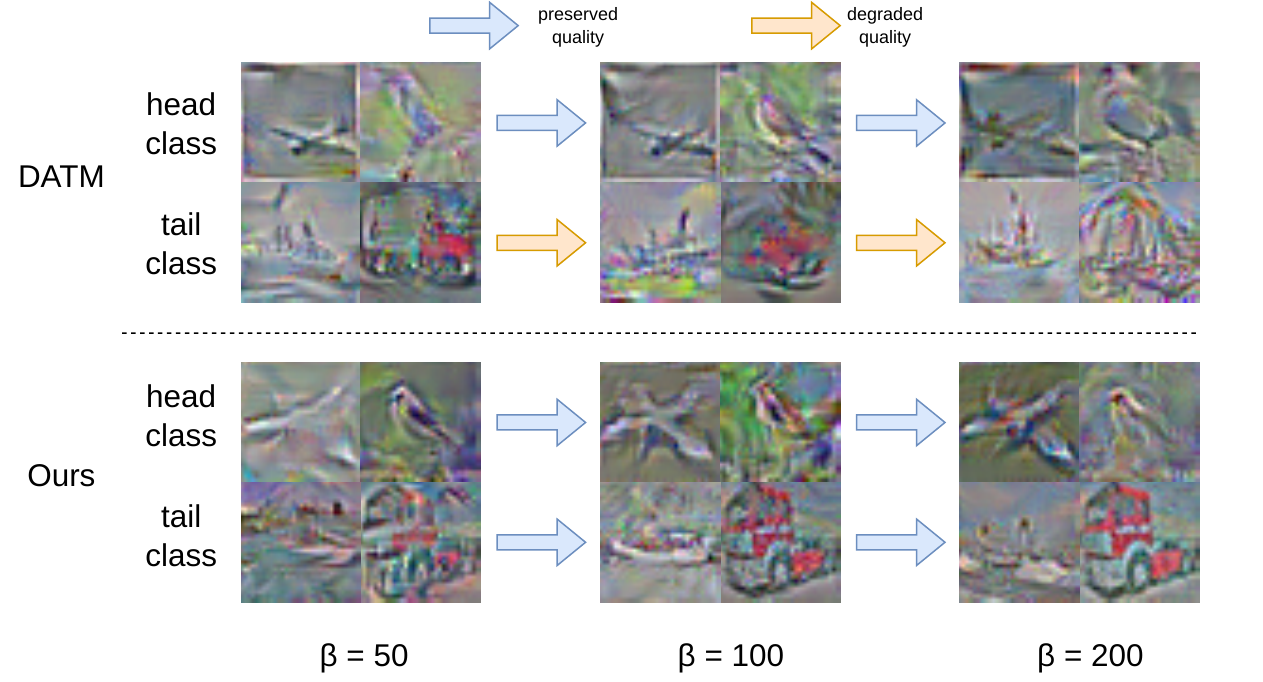}
    \caption{\textbf{Visualization of distilled datasets.} We visualize the images from the distilled dataset for DATM and our method. We can observe that as the imbalance factor increases, images from DATM preserve the quality on head classes, but degrade on tail classes. On the contrary, our method is able to preserve good quality in all classes.}
    \label{fig:vis_comparison}
\end{figure*}

The distilled dataset is visualized in Figure~\ref{fig:vis_comparison}. We visualized the synthetic dataset images of DATM and our method under three different imbalance factors, $\beta=50$, $\beta=100$, $\beta=200$. We observe in the figure that: As the imbalance factor increases, the DATAM distilled image quality degrades and contains more noise and distortions. Instead, our distilled dataset can still maintain good quality.

\begin{table*}[t]
    \centering
    \resizebox{0.8\linewidth}{!}{
    \begin{tabular}{c|cc|cc|ccc|ccc|cc}
        \toprule
        Dataset & $\beta$ & IPC & $N_{rep}$ & $N_{cls}$ & $T^{-}_{rep}$ & $T_{rep}$ & $T^{+}_{rep}$ & $T^{-}_{cls}$ & $T_{cls}$ & $T^{+}_{cls}$ & $\lambda_{rep}$ & $\lambda_{cls}$ \\
        \midrule
        \multirow{12}{*}{CIFAR-10-LT} & \multirow{3}{*}{10} & 10 & 80 & 80 & 0 & 10 & 20 & 0 & 1 & 1 & 1.0 & 0.2  \\
        && 20 & 80 & 80 & 0 & 10 & 20 & 0 & 1 & 1 & 1.0 & 0.2  \\
        && 50 & 80 & 80 & 0 & 20 & 40 & 0 & 1 & 1 & 1.0 & 0.2 \\ 
         & \multirow{3}{*}{50} & 10 & 80 & 80 & 0 & 10 & 20 & 0 & 1 & 1 & 0.5 & 0.5 \\
        && 20 & 80 & 80 & 0 & 10 & 20 & 0 & 1 & 1 & 0.5 & 0.5 \\
        && 50 & 80 & 80 & 0 & 20 & 40 & 0 & 1 & 1 & 0.5 & 0.5 \\ 
         & \multirow{3}{*}{100} & 10 & 80 & 80 & 0 & 10 & 20 & 0 & 1 & 1 & 1.0 & 0.5 \\
        && 20 & 80 & 80 & 0 & 10 & 20 & 0 & 1 & 1 & 1.0 & 0.5 \\
        && 50 & 80 & 80 & 0 & 20 & 40 & 0 & 1 & 1 & 1.0 & 0.5 \\ 
         & \multirow{3}{*}{200} & 10 & 80 & 80 & 0 & 10 & 20 & 0 & 1 & 1 & 0.1 & 1.0 \\
        && 20 & 80 & 80 & 0 & 10 & 20 & 0 & 1 & 1 & 0.1 & 1.0 \\
        && 50 & 80 & 80 & 0 & 20 & 40 & 0 & 1 & 1 & 0.1 & 1.0 \\ \midrule

        \multirow{6}{*}{CIFAR-100-LT} & \multirow{3}{*}{10} & 10 & 40 & 20 & 0 & 30 & 50 & 0 & 1 & 1 & 1.0 & 0.1 \\
        && 20 & 40 & 20 & 20 & 70 & 70 & 0 & 1 & 1 & 1.0 & 0.1 \\
        && 50 & 40 & 20 & 20 & 70 & 70 & 0 & 1 & 1 & 1.0 & 0.1 \\ 
         & \multirow{3}{*}{20} & 10 & 40 & 20 & 0 & 30 & 50 & 0 & 1 & 1 & 1.0 & 0.1 \\
        && 20 & 40 & 20 & 20 & 70 & 70 & 0 & 1 & 1 & 1.0 & 0.1 \\
        && 50 & 40 & 20 & 20 & 70 & 70 & 0 & 1 & 1 & 1.0 & 0.1 \\ 
        
        \bottomrule
    \end{tabular}
    }
    \caption{\textbf{Hyper-parameters for different settings.}}
    \label{tab:hyper}
\end{table*}


\label{sec:limitations}

%% file: main.bbl
\begin{thebibliography}{50}
\providecommand{\natexlab}[1]{#1}
\providecommand{\url}[1]{\texttt{#1}}
\expandafter\ifx\csname urlstyle\endcsname\relax
  \providecommand{\doi}[1]{doi: #1}\else
  \providecommand{\doi}{doi: \begingroup \urlstyle{rm}\Url}\fi

\bibitem[Alshammari et~al.(2022{\natexlab{a}})Alshammari, Wang, Ramanan, and Kong]{ltr_regularization}
Shaden Alshammari, Yuxiong Wang, Deva Ramanan, and Shu Kong.
\newblock Long-tailed recognition via weight balancing.
\newblock In \emph{CVPR}, 2022{\natexlab{a}}.

\bibitem[Alshammari et~al.(2022{\natexlab{b}})Alshammari, Wang, Ramanan, and Kong]{alshammari2022longweightbalancing}
Shaden Alshammari, Yu-Xiong Wang, Deva Ramanan, and Shu Kong.
\newblock Long-tailed recognition via weight balancing.
\newblock In \emph{CVPR}, 2022{\natexlab{b}}.

\bibitem[Buda et~al.(2018)Buda, Maki, and Mazurowski]{buda2018systematic_under1}
Mateusz Buda, Atsuto Maki, and Maciej~A Mazurowski.
\newblock A systematic study of the class imbalance problem in convolutional neural networks.
\newblock \emph{Neural networks}, 2018.

\bibitem[Cazenavette et~al.(2022)Cazenavette, Wang, Torralba, Efros, and Zhu]{cazenavette2022datasetmtt}
George Cazenavette, Tongzhou Wang, Antonio Torralba, Alexei~A Efros, and Jun-Yan Zhu.
\newblock Dataset distillation by matching training trajectories.
\newblock In \emph{CVPR}, 2022.

\bibitem[Chou et~al.(2020)Chou, Chang, Pan, Wei, and Juan]{chou2020remix_aug2}
Hsin-Ping Chou, Shih-Chieh Chang, Jia-Yu Pan, Wei Wei, and Da-Cheng Juan.
\newblock Remix: rebalanced mixup.
\newblock In \emph{Computer Vision--ECCV 2020 Workshops: Glasgow, UK, August 23--28, 2020, Proceedings, Part VI 16}, 2020.

\bibitem[Cui et~al.(2023)Cui, Wang, Si, and Hsieh]{cui2023scalingtesla}
Justin Cui, Ruochen Wang, Si Si, and Cho-Jui Hsieh.
\newblock Scaling up dataset distillation to imagenet-1k with constant memory.
\newblock In \emph{ICML}, 2023.

\bibitem[Cui et~al.(2019)Cui, Jia, Lin, Song, and Belongie]{cui2019classcifarlt}
Yin Cui, Menglin Jia, Tsung-Yi Lin, Yang Song, and Serge Belongie.
\newblock Class-balanced loss based on effective number of samples.
\newblock In \emph{CVPR}, 2019.

\bibitem[Deng et~al.(2009)Deng, Dong, Socher, Li, Li, and Fei-Fei]{deng2009imagenet}
Jia Deng, Wei Dong, Richard Socher, Li-Jia Li, Kai Li, and Li Fei-Fei.
\newblock Imagenet: A large-scale hierarchical image database.
\newblock In \emph{CVPR}, 2009.

\bibitem[Du et~al.(2023{\natexlab{a}})Du, Yang, Jia, Nan, Chen, and Yang]{du2023globalGLMC}
Fei Du, Peng Yang, Qi Jia, Fengtao Nan, Xiaoting Chen, and Yun Yang.
\newblock Global and local mixture consistency cumulative learning for long-tailed visual recognitions.
\newblock In \emph{CVPR}, 2023{\natexlab{a}}.

\bibitem[Du et~al.(2023{\natexlab{b}})Du, Jiang, Tan, Zhou, and Li]{du2023minimizing}
Jiawei Du, Yidi Jiang, Vincent~YF Tan, Joey~Tianyi Zhou, and Haizhou Li.
\newblock Minimizing the accumulated trajectory error to improve dataset distillation.
\newblock In \emph{CVPR}, 2023{\natexlab{b}}.

\bibitem[Gidaris and Komodakis(2018)]{gidaris2018dynamic_aug1}
Spyros Gidaris and Nikos Komodakis.
\newblock Dynamic few-shot visual learning without forgetting.
\newblock In \emph{CVPR}, 2018.

\bibitem[Gu et~al.(2024)Gu, Vahidian, Kungurtsev, Wang, Jiang, You, and Chen]{gu2024efficientminimax}
Jianyang Gu, Saeed Vahidian, Vyacheslav Kungurtsev, Haonan Wang, Wei Jiang, Yang You, and Yiran Chen.
\newblock Efficient dataset distillation via minimax diffusion.
\newblock In \emph{CVPR}, 2024.

\bibitem[Guo et~al.(2024)Guo, Wang, Cazenavette, Li, Zhang, and You]{datm}
Ziyao Guo, Kai Wang, George Cazenavette, Hui Li, Kaipeng Zhang, and Yang You.
\newblock Towards lossless dataset distillation via difficulty-aligned trajectory matching.
\newblock In \emph{ICLR}, 2024.

\bibitem[Haixiang et~al.(2017)Haixiang, Yijing, Shang, Mingyun, Yuanyue, and Bing]{haixiang2017learning_oversample1}
Guo Haixiang, Li Yijing, Jennifer Shang, Gu Mingyun, Huang Yuanyue, and Gong Bing.
\newblock Learning from class-imbalanced data: Review of methods and applications.
\newblock \emph{Expert systems with applications}, 2017.

\bibitem[Hinton et~al.(2015)Hinton, Vinyals, and Dean]{kd}
Geoffrey Hinton, Oriol Vinyals, and Jeff Dean.
\newblock Distilling the knowledge in a neural network.
\newblock \emph{arXiv preprint arXiv:1503.02531}, 2015.

\bibitem[Hinton et~al.(2012)Hinton, Srivastava, Krizhevsky, Sutskever, and Salakhutdinov]{hinton2012improvingmaxnrom1}
Geoffrey~E Hinton, Nitish Srivastava, Alex Krizhevsky, Ilya Sutskever, and Ruslan~R Salakhutdinov.
\newblock Improving neural networks by preventing co-adaptation of feature detectors.
\newblock \emph{arXiv preprint arXiv:1207.0580}, 2012.

\bibitem[Iyer et~al.(2021)Iyer, Khargoankar, Bilmes, and Asanani]{iyer2021submodulargraphcut}
Rishabh Iyer, Ninad Khargoankar, Jeff Bilmes, and Himanshu Asanani.
\newblock Submodular combinatorial information measures with applications in machine learning.
\newblock In \emph{Algorithmic Learning Theory}, 2021.

\bibitem[Janowczyk and Madabhushi(2016)]{janowczyk2016deep_oversample3}
Andrew Janowczyk and Anant Madabhushi.
\newblock Deep learning for digital pathology image analysis: A comprehensive tutorial with selected use cases.
\newblock \emph{Journal of pathology informatics}, 2016.

\bibitem[Ju et~al.(2021)Ju, Wang, Wang, Liu, Zhao, Drummond, Mahapatra, and Ge]{medical_lt}
Lie Ju, Xin Wang, Lin Wang, Tongliang Liu, Xin Zhao, Tom Drummond, Dwarikanath Mahapatra, and Zongyuan Ge.
\newblock Relational subsets knowledge distillation for long-tailed retinal diseases recognition.
\newblock In \emph{MICCAI}, 2021.

\bibitem[Kang et~al.(2020)Kang, Xie, Rohrbach, Yan, Gordo, Feng, and Kalantidis]{kang2019decoupling}
Bingyi Kang, Saining Xie, Marcus Rohrbach, Zhicheng Yan, Albert Gordo, Jiashi Feng, and Yannis Kalantidis.
\newblock Decoupling representation and classifier for long-tailed recognition.
\newblock In \emph{ICLR}, 2020.

\bibitem[Kim et~al.(2022)Kim, Kim, Oh, Yun, Song, Jeong, Ha, and Song]{idc}
Jang-Hyun Kim, Jinuk Kim, Seong~Joon Oh, Sangdoo Yun, Hwanjun Song, Joonhyun Jeong, Jung-Woo Ha, and Hyun~Oh Song.
\newblock Dataset condensation via efficient synthetic-data parameterization.
\newblock In \emph{ICML}, 2022.

\bibitem[Krizhevsky(2012)]{cifar}
Alex Krizhevsky.
\newblock Learning multiple layers of features from tiny images.
\newblock \emph{University of Toronto}, 2012.

\bibitem[Krizhevsky et~al.(2017)Krizhevsky, Sutskever, and Hinton]{krizhevsky2017imagenetconvnet}
Alex Krizhevsky, Ilya Sutskever, and Geoffrey~E Hinton.
\newblock Imagenet classification with deep convolutional neural networks.
\newblock \emph{Communications of the ACM}, 2017.

\bibitem[Le and Yang(2015)]{tiny-imagenet}
Yann Le and Xuan Yang.
\newblock Tiny imagenet visual recognition challenge.
\newblock \emph{CS 231N}, 2015.

\bibitem[Li et~al.(2024)Li, Zhong, Liang, Zhou, Shi, Wang, Zhao, Zhao, Wang, Qin, Liu, Zhang, Zhou, Zhu, Wang, Li, Zhang, Liu, Huang, Lyu, Lv, Jin, Akata, Gu, Vedantam, Shou, Deng, Yan, Shang, Cazenavette, Wu, Cui, Chen, Yao, Kellis, Plataniotis, Zhao, Wang, You, and Wang]{li2024ddranking}
Zekai Li, Xinhao Zhong, Zhiyuan Liang, Yuhao Zhou, Mingjia Shi, Ziqiao Wang, Wangbo Zhao, Xuanlei Zhao, Haonan Wang, Ziheng Qin, Dai Liu, Kaipeng Zhang, Tianyi Zhou, Zheng Zhu, Kun Wang, Guang Li, Junhao Zhang, Jiawei Liu, Yiran Huang, Lingjuan Lyu, Jiancheng Lv, Yaochu Jin, Zeynep Akata, Jindong Gu, Rama Vedantam, Mike Shou, Zhiwei Deng, Yan Yan, Yuzhang Shang, George Cazenavette, Xindi Wu, Justin Cui, Tianlong Chen, Angela Yao, Manolis Kellis, Konstantinos~N. Plataniotis, Bo Zhao, Zhangyang Wang, Yang You, and Kai Wang.
\newblock Dd-ranking: Rethinking the evaluation of dataset distillation.
\newblock GitHub repository, 2024.

\bibitem[Lin et~al.(2017)Lin, Goyal, Girshick, He, and Doll{\'a}r]{focal_loss}
Tsung-Yi Lin, Priya Goyal, Ross Girshick, Kaiming He, and Piotr Doll{\'a}r.
\newblock Focal loss for dense object detection.
\newblock In \emph{ICCV}, 2017.

\bibitem[Liu et~al.(2018)Liu, Simonyan, and Yang]{darts}
Hanxiao Liu, Karen Simonyan, and Yiming Yang.
\newblock Darts: Differentiable architecture search.
\newblock \emph{arXiv preprint arXiv:1806.09055}, 2018.

\bibitem[Liu et~al.(2023)Liu, Gu, Wang, Zhu, Jiang, and You]{dream}
Yanqing Liu, Jianyang Gu, Kai Wang, Zheng Zhu, Wei Jiang, and Yang You.
\newblock Dream: Efficient dataset distillation by representative matching.
\newblock In \emph{ICCV}, 2023.

\bibitem[Liu et~al.(2019)Liu, Miao, Zhan, Wang, Gong, and Yu]{liu2019largeimagenetlt}
Ziwei Liu, Zhongqi Miao, Xiaohang Zhan, Jiayun Wang, Boqing Gong, and Stella~X Yu.
\newblock Large-scale long-tailed recognition in an open world.
\newblock In \emph{CVPR}, 2019.

\bibitem[Loo et~al.(2022)Loo, Hasani, Amini, and Rus]{loo2022efficientkernel1}
Noel Loo, Ramin Hasani, Alexander Amini, and Daniela Rus.
\newblock Efficient dataset distillation using random feature approximation.
\newblock In \emph{NeurIPS}, 2022.

\bibitem[Masarczyk and Tautkute(2020)]{continual}
Wojciech Masarczyk and Ivona Tautkute.
\newblock Reducing catastrophic forgetting with learning on synthetic data.
\newblock In \emph{Proceedings of the IEEE/CVF Conference on Computer Vision and Pattern Recognition Workshops}, 2020.

\bibitem[Nguyen et~al.(2020)Nguyen, Chen, and Lee]{nguyen2020datasetkernel3}
Timothy Nguyen, Zhourong Chen, and Jaehoon Lee.
\newblock Dataset meta-learning from kernel ridge-regression.
\newblock \emph{arXiv preprint arXiv:2011.00050}, 2020.

\bibitem[Peng et~al.(2020)Peng, Bu, Sun, Zhang, Tan, and Yan]{peng2020large_oversample2}
Junran Peng, Xingyuan Bu, Ming Sun, Zhaoxiang Zhang, Tieniu Tan, and Junjie Yan.
\newblock Large-scale object detection in the wild from imbalanced multi-labels.
\newblock In \emph{CVPR}, 2020.

\bibitem[Ren et~al.(2020)Ren, Yu, Ma, Zhao, Yi, et~al.]{ren2020balancedsoftmaxloss}
Jiawei Ren, Cunjun Yu, Xiao Ma, Haiyu Zhao, Shuai Yi, et~al.
\newblock Balanced meta-softmax for long-tailed visual recognition.
\newblock In \emph{NeurIPS}, 2020.

\bibitem[Sachdeva and McAuley(2023)]{dd_survey}
Noveen Sachdeva and Julian McAuley.
\newblock Data distillation: A survey.
\newblock \emph{arXiv preprint arXiv:2301.04272}, 2023.

\bibitem[Sener and Savarese(2017)]{sener2017activekcenter}
Ozan Sener and Silvio Savarese.
\newblock Active learning for convolutional neural networks: A core-set approach.
\newblock \emph{arXiv preprint arXiv:1708.00489}, 2017.

\bibitem[Shalev-Shwartz et~al.(2007)Shalev-Shwartz, Singer, and Srebro]{shalev2007pegasosmaxnorm2}
Shai Shalev-Shwartz, Yoram Singer, and Nathan Srebro.
\newblock Pegasos: Primal estimated sub-gradient solver for svm.
\newblock In \emph{ICML}, 2007.

\bibitem[Shang et~al.(2024)Shang, Yuan, and Yan]{shang2024mim4dd}
Yuzhang Shang, Zhihang Yuan, and Yan Yan.
\newblock Mim4dd: Mutual information maximization for dataset distillation.
\newblock In \emph{NeurIPS}, 2024.

\bibitem[Shao et~al.(2024)Shao, Yin, Zhou, Zhang, and Shen]{shao2024generalized}
Shitong Shao, Zeyuan Yin, Muxin Zhou, Xindong Zhang, and Zhiqiang Shen.
\newblock Generalized large-scale data condensation via various backbone and statistical matching.
\newblock In \emph{CVPR}, 2024.

\bibitem[Su et~al.(2024)Su, Hou, Gao, Tian, and Tang]{su2024d4m}
Duo Su, Junjie Hou, Weizhi Gao, Yingjie Tian, and Bowen Tang.
\newblock D\^{} 4: Dataset distillation via disentangled diffusion model.
\newblock In \emph{CVPR}, 2024.

\bibitem[Wang et~al.(2022)Wang, Zhao, Peng, Zhu, Yang, Wang, Huang, Bilen, Wang, and You]{cafe}
Kai Wang, Bo Zhao, Xiangyu Peng, Zheng Zhu, Shuo Yang, Shuo Wang, Guan Huang, Hakan Bilen, Xinchao Wang, and Yang You.
\newblock Cafe: Learning to condense dataset by aligning features.
\newblock In \emph{CVPR}, 2022.

\bibitem[Wang et~al.(2024)Wang, Li, Cheng, Khaki, Sajedi, Vedantam, Plataniotis, Hauptmann, and You]{wang2024emphasizing}
Kai Wang, Zekai Li, Zhi-Qi Cheng, Samir Khaki, Ahmad Sajedi, Ramakrishna Vedantam, Konstantinos~N Plataniotis, Alexander Hauptmann, and Yang You.
\newblock Emphasizing discriminative features for dataset distillation in complex scenarios.
\newblock \emph{arXiv preprint arXiv:2410.17193}, 2024.

\bibitem[Wang et~al.(2018)Wang, Zhu, Torralba, and Efros]{wang2018dataset}
Tongzhou Wang, Jun-Yan Zhu, Antonio Torralba, and Alexei~A Efros.
\newblock Dataset distillation.
\newblock \emph{arXiv preprint arXiv:1811.10959}, 2018.

\bibitem[Yin et~al.(2023)Yin, Xing, and Shen]{yin2023squeeze}
Zeyuan Yin, Eric Xing, and Zhiqiang Shen.
\newblock Squeeze, recover and relabel: Dataset condensation at imagenet scale from a new perspective.
\newblock In \emph{NeurIPS}, 2023.

\bibitem[Zhang et~al.(2023)Zhang, Kang, Hooi, Yan, and Feng]{lt_survey}
Yifan Zhang, Bingyi Kang, Bryan Hooi, Shuicheng Yan, and Jiashi Feng.
\newblock Deep long-tailed learning: A survey.
\newblock \emph{IEEE transactions on pattern analysis and machine intelligence}, 2023.

\bibitem[Zhao et~al.(2020)Zhao, Mopuri, and Bilen]{zhao2020datasetdc}
Bo Zhao, Konda~Reddy Mopuri, and Hakan Bilen.
\newblock Dataset condensation with gradient matching.
\newblock \emph{arXiv preprint arXiv:2006.05929}, 2020.

\bibitem[Zhao et~al.(2023)Zhao, Li, Qin, and Yu]{zhao2023improvedidm}
Ganlong Zhao, Guanbin Li, Yipeng Qin, and Yizhou Yu.
\newblock Improved distribution matching for dataset condensation.
\newblock In \emph{CVPR}, 2023.

\bibitem[Zhao et~al.(2024)Zhao, Shang, Wu, and Yan]{dqas}
Zhenghao Zhao, Yuzhang Shang, Junyi Wu, and Yan Yan.
\newblock Dataset quantization with active learning based adaptive sampling.
\newblock In \emph{ECCV}, 2024.

\bibitem[Zhong et~al.(2021)Zhong, Cui, Liu, and Jia]{ltr_calibration}
Zhisheng Zhong, Jiequan Cui, Shu Liu, and Jiaya Jia.
\newblock Improving calibration for long-tailed recognition.
\newblock In \emph{CVPR}, 2021.

\bibitem[Zhou et~al.(2022)Zhou, Nezhadarya, and Ba]{zhou2022datasetkernel2}
Yongchao Zhou, Ehsan Nezhadarya, and Jimmy Ba.
\newblock Dataset distillation using neural feature regression.
\newblock In \emph{NeurIPS}, 2022.

\end{thebibliography}
